%% file: main_acl_submission.tex
\pdfoutput=1
\documentclass[11pt]{article}

\usepackage[final]{acl}

\usepackage{array}
\usepackage{times}
\usepackage{latexsym}
\usepackage{graphicx}
\usepackage{booktabs} 

\usepackage{enumitem}
\setitemize{leftmargin=*}

\usepackage[T1]{fontenc}
\usepackage[utf8]{inputenc}

\usepackage{microtype}

\usepackage{inconsolata}

\title{Controlled Automatic Task-Specific Synthetic Data Generation for Hallucination Detection}

\author{Yong Xie \and Karan Aggarwal \and Aitzaz Ahmad \and Stephen Lau \\
        Amazon \\ yonxie, kagg, aitzaza, lausteph@amazon.com}

\begin{document}
\maketitle
\begin{abstract}
We present a novel approach to automatically generate non-trivial task-specific synthetic datasets for hallucination detection. Our approach features a two-step generation-selection pipeline, using hallucination pattern guidance and a language style alignment during generation. Hallucination pattern guidance leverages the most important task-specific hallucination patterns while language style alignment aligns the style of the synthetic dataset with benchmark text. To obtain robust supervised detectors from synthetic datasets, we also adopt a data mixture strategy to improve performance robustness and generalization. 
Our results on three datasets show that our generated hallucination text is more closely aligned with non-hallucinated text versus baselines, to train hallucination detectors with better generalization.
Our hallucination detectors trained on synthetic datasets outperform in-context-learning (ICL)-based detectors by a large margin of 32\%. Our extensive experiments confirm the benefits of our approach with cross-task and cross-generator generalization. Our data-mixture-based training further improves the generalization and robustness of hallucination detection.
\end{abstract}

\section{Introduction}

The ability of large language models (LLMs) to generate human-like text \citep{llama2, instructGPT, falcon} has advanced significantly in recent years, enabling a wide range of applications, from document summarizers \citep{jin2024comprehensive, lala2023paperqa} to coding assistants \citep{yeticstiren2023evaluating}. However, one of the key challenges in deploying these models is the risk of hallucinations — the generation of plausible but factually incorrect information. Hallucinations can occur when the model makes up details that are not grounded in the input or the given context, leading to the generation of misinformation or nonsensical outputs. The tendency to hallucinate raises concerns about the safety and reliability of LLMs in critical domains like finance 
\citep{continual_pretraining}
and healthcare \citep{med-palm2}.

Despite the debate on the categorization of various types of hallucinations~\citep{hall_survey_huang2023, zhang2023siren}, the type of hallucination is task-dependent.
For example, we are likely to see more factual hallucinations in an open-ended question-answering scenario while more logical hallucinations in code generation tasks~\citep{liu2024exploring}.
%The importance attributed to each hallucination type is again task-dependent, depending on the nature of the task.
% For instance, hallucinations producing factually incorrect answers can erase a customer's trust in chatbot-style AI assistants.
Therefore, it is critical to customize the hallucination evaluation and detection to the specific task.

\textit{Post-hoc} hallucination detection approaches detect hallucinations once they have been generated by the LLM. %Hallucination detectors are typically built by training a classifier on observed hallucinations or on open-sourced hallucination datasets. 
However, the challenge lies in obtaining  hallucination datasets for training hallucination detectors in the absence of such hallucinated data. 
Synthetic hallucination datasets are one of the most commonly used methods to build such detectors~\cite{HaluEval_LiCZNW23}. While these detectors provide a high accuracy, critical aspects on the usability of such detectors are overlooked
: \textit{1) can these detectors capture the task-specific hallucination patterns beyond benchmark datasets? 2) how versatile are these detectors in the environment of fast LLM iteration and application development, despite developed for being task-specific?}

% do these detectors capture trivial language features such as style to distinguish between non-hallucinated and synthetically generated hallucinated text?  }

% in order to build such detectors, we need access to observed hallucinations, which are harder to get before the solution is actually being used. Since hallucinations are task-specific, detectors trained on open-sourced datasets may not be relevant. Hence, there is a need for task-specific hallucination detectors pre-production to ensure a trustworthy solution for users.
%is an important approach to mitigate the adversarial effects of hallucination. Despite the fact that hallucination detectors can be obtained through supervised training on open-ended hallucination datasets, The resulting models cannot guarantee acceptable performance on the target application as hallucinations are highly task-specific. In the context of application development, it is also undesirable to curate hallucination datasets by collecting observed hallucinations in the wild; the existence of hallucinations erodes customer trust.

\begin{figure*}[t]
    \centering
    \includegraphics[width=0.8\textwidth]{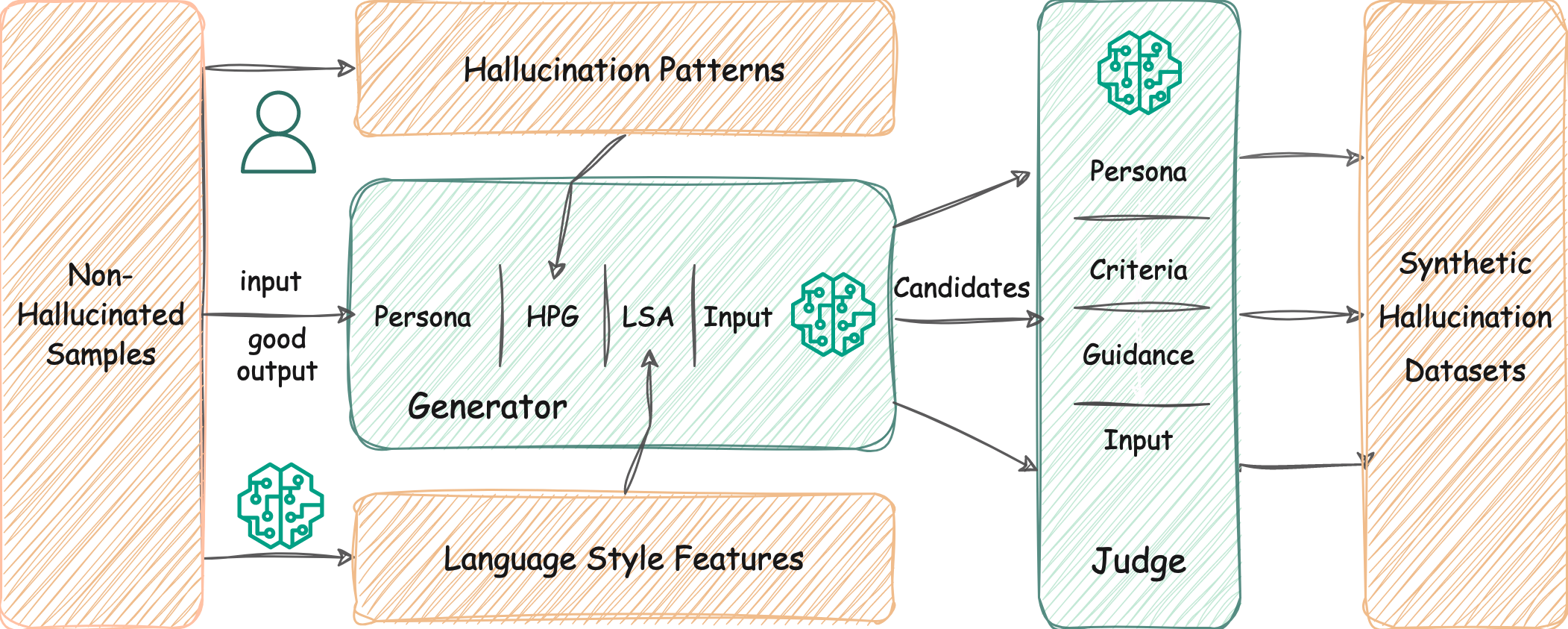}
    \caption{Automatic generation pipeline. We use non-hallucinated samples to generate the synthetic hallucination dataset with two inputs to the generator: human defined Hallucination Patterns and Language Style Features with language style features, like text tone. These are used by the Generator LLM to generate hallucinated samples, which are then judged by a LLM Judge to finally select the most plausible hallucinated samples.}
    \label{fig:enter-label}
    \vspace{-15pt}
\end{figure*}

In this paper, we propose a generic approach to curate synthetic datasets for training hallucination detectors (hallucination datasets), as shown in Figure~\ref{fig:enter-label}. Our approach features a two-step \textit{Generation-Selection} pipeline: we first generate a group of hallucinated candidates for a given input through an LLM (generator) and then select the best candidate through an LLM (judge) based on a given criteria. The synthetic datasets are then used as training data to develop \textit{post-hoc} hallucination detectors. To answer the two key questions, we propose two design features in the generation step to customize hallucination generation and improve dataset quality: \textit{Hallucination Pattern Guidance (HPG)} and \textit{Language Style Alignment (LSA)}.

% Motivated by the first questions, we propose a generic approach to curate synthetic datasets for training hallucination detectors (hallucination datasets), as shown in Figure~\ref{fig:enter-label}, 

% to align non-hallucinated and synthetically generated hallucinated text to create non-trivial hallucination detectors.

%Our work answers the practical demand for task-specific hallucination detectors.
% Our approach features a two-step \textit{Generation-Selection} pipeline: we first generate a group of hallucinated candidates for a given input through an LLM (generator) and then select the best candidate through an LLM (judge) based on a given criteria. Moreover, 

Hallucination pattern guidance is motivated by the need for task-specific hallucinated samples. Starting from a set of pre-defined hallucination patterns using a little human effort, we prompt the generator to generate hallucinated samples conforming to the given patterns. This design enables the synthetic datasets to cover both general and task-specific hallucination patterns. It tailors the generation pipeline to the concerning and consequential hallucinating behaviors observed in practice.

Language style alignment is designed to improve detector generalization by improving data quality and mitigating detector shortcuts. 
% Shortcuts picked up by detectors during training negatively impact generalization by misleading the detectors to focus on superficial features rather than hallucination patterns. 
Language style discrepancy between hallucinated and non-hallucinated data is empirically observed, which can be easily picked up during training and thus erodes the detector generalization.  
LSA mitigates the issue by aligning the text characteristics of hallucinated samples with those of non-hallucinated LLM responses. %Recognizing that analyzing language style requires expertise in linguistics, 
We propose a hierarchical \textit{Language Style Discovery} algorithm by leveraging LLMs to distill the styles into a small feature set used as guidelines to govern hallucination generation. This refinement aligns generations with non-hallucinated text, generates more challenging training data, and leads to detectors with better generalization. 
% This refinement helps align generations with non-hallucinated text, \emph{making our hallucinated dataset more challenging as it becomes harder to distinguish between hallucinated and non-hallucinated text based off trivial aspects such as language style.} %Hence, our hallucination detectors are more powerful.

%We fine-tune RoBERTa \citep{roberta} on the synthetic hallucination datasets to develop supervised hallucination detectors.
We conduct experiments on three conversational benchmarks by generating synthetic hallucination datasets to train hallucination detectors. 
Our hallucination detector achieves an F1 score of 0.938 on average over three benchmarks and six different generators, outperforming the in-context learning based LLM detectors by a large margin of 32.5\%. It implies that dedicated detectors trained on synthetic datasets are powerful and cost-effective options for \textit{post-hoc} hallucination detection.

To answer question of generalization, we show that \emph{our synthetic hallucinations are more similar to non-hallucinated samples} through various text distance metrics compared to existing approaches~\citep{HaluEval_LiCZNW23, YuZZMRKSZ23}. 
% showing the effectiveness of our generation pipeline.
Moreover, our empirical investigation based off detector performance documents strong generalization capability demonstrated by our pipeline and detectors in three dimensions: \textit{1) Out-of-generator generalization—detectors trained on a dataset generated by one LLM can be used on generations of other LLMs; 2) out-of-pattern generalization—detectors trained on a hallucination pattern(s) can generalize on unseen patterns; 3) out-of-task generalization—detectors trained on one task can generalize on other tasks.} In all three scenarios, the supervised detectors trained with synthetic datasets generated by our pipeline deliver superior performance than ICL detectors and better generalization than baselines. The results confirm that detectors trained with our pipeline are also versatile, despite being developed for being task specific.

To summarize, we make the following contributions:

%it is not always feasible to generate synthetic datasets with the same production LLMs in commercial use cases. For example, one may leverage the superior performance of Claude 3, but any data generated by the model is restricted to being used for third-party model development by the model license.
% We investigate this cross-model generalization capability %by evaluating hallucination detectors on synthetic datasets generated by various LLMs (out-of-generator datasets).
% with our results revealing a generalization issue: hallucination detectors perform worse on out-of-generator datasets than in-generator dataset(s). %Moreover, the generalization ability varies across different models, suggesting that detector performance is less robust and foreseeable when deployed in out-of-generator applications.
% To mitigate the generalization problem, we propose a simple but effective data mixture strategy to obtain a more diverse training corpus by merging synthetic datasets generated by multiple LLMs. Our experiments demonstrate that the strategy improves generalization performance and robustness consistently.

% \vspace{-5pt}
\begin{itemize}
\item We propose a quality data generating pipeline that enables hallucination pattern customization; 
\item we design a novel approach to improve data quality by aligning the synthetic hallucinations with the non-hallucinated text's language style;
\item through extensive experiments, we empirically document that supervised detectors consistently exhibit cross-task, cross-generator and cross-pattern generalization capabilities.  
% and
% \item We introduce a more generalized and robust hallucination detection strategy using a mixture of synthetic data generation through multiple LLMs.
\end{itemize}

%We propose two techniques to mitigate the generalization problem and increase the performance robustness. First, we utilize a \textit{Language Feature Discovery} module to automatically identify the language style features (such as writing style, lengths, tone, etc.) in non-hallucinated data, and then prompt the LLM generators to mimic the discovered language style features when generating hallucinated datasets. Second, we increase the diversity of training corpus through \textit{Data Mixture}, i.e., synthetic datasets produced by various LLMs are mixed together to form a training dataset for supervised detectors. Our ablation study show that the two techniques mitigate the generalization problem and reduces the performance variability on out-of-generator datasets. 

\vspace{-6pt}
\section{Methodology}
% \vspace{-5pt}
\paragraph{Problem Setting.} 
% In this work, we focus on settings where the hallucinations have not been observed, i.e., before the application has been put into production. The only thing we assume is the presence of a {benchmark dataset}, which is a set of non-hallucinated input-output pairs either from humans or from an LLM to build our hallucination generation pipeline. 
In this work, we assume that there exists a {benchmark dataset}, which is a set of non-hallucinated input-output pairs either from humans or from an LLM, similar to HaluEval~\cite{HaluEval_LiCZNW23}. This benchmark is used to build our hallucination generation pipeline. 
We rely on human judgment to provide hallucination patterns that need to be detected. Our objective is to create a synthetic dataset that contains both hallucinated and non-hallucinated input-output pairs. Hallucination detection is formulated as a binary classification task: given an input and an LLM output, a detector determines whether the LLM output is hallucinated with respect to the input.

Our proposed approach features an automatic \textit{Generation-Selection} pipeline with \textit{Hallucination Pattern Guidance} (HPG) and \textit{Language Style Alignment} (LSA). The generation-selection mechanism consists of a generation step to obtain a set of candidate hallucinated samples and a selection step to pick the most plausible one, which ensures the generation quality. HPG and LSA are two versatile modules integrated into the generation step. 
% The former guides the generator to produce task-specific synthetic datasets covering the most relevant hallucination patterns for target tasks, and the latter aligns the generations with non-hallucinated text in order to improve data quality.
%We lay out the design details and discuss the motivation in this section.

\subsection{Generation-Selection Pipeline}

While using synthetic data for hallucinations is becoming more common, the quality of the hallucination data can be low, especially for automatic approaches without human intervention. To resolve the issue, we adopt the two-step \textit{first-generate-then-select} design \citep{HaluEval_LiCZNW23} to ensure generation quality. Specifically, two LLMs (not necessarily the same) act in the roles of generator and judge separately. The generator produces a set of hallucinated outputs per input according to predetermined patterns. The judge scores the hallucinated candidates by given criteria, and the one with the highest score is selected. This step improves generation quality by selecting the best among the group of candidates.

\vspace{-5pt}
\paragraph{\textbf{Generator.}} Generator is a prompted LLM performing the task of hallucinated sample generation. We utilize LLMs to generate hallucinated data, as they are proven to generate high-quality text while following instructions. The key here is to properly design the prompts for hallucination guidance and language style alignment. Besides, it is important to carefully specify the persona in the system prompt to work around the safety policies in place that prevent LLMs from generating hallucinations. We adopt the chain-of-thought (CoT) \citep{cot_wei} prompt and ask the generator to provide rationale for the generated samples, inspired by \citet{LLM_generation_attribute_manipulation}. Our generator prompt is structured as follows: The prompt starts with a definition of persona customized for target tasks, which is followed by a section of HPG consisting of the pattern description and one demonstration example. The next is the LSA section, which comprises itemized guidelines for text generation. The prompt ends with an input and brief instructions on the output format. The details of prompts are deferred to Section \ref{sec: hallucination_pattern_guidance} and Section \ref{sec: language_feature_alignment}.

% \vspace{-5pt} 
\paragraph{\textbf{Judge.}} A judge is a prompted LLM performing the task of evaluating the quality of hallucination candidates according to the given criteria. We prompt the judge to score the candidates on a scale of 1 to 10, and the candidate with the highest score is then selected as the hallucinated output for the given input. We follow this scoring mechanism instead of directly selecting the best candidate out of the generated candidates because the scoring approach is less prone to LLMs' positional bias \citep{llm_not_fair_evaluators}. Moreover, we also adopt the CoT prompt to generate rationale for the scores to improve accuracy. The judge prompt (see Appendix~\ref{sec:prompt}) consists of a customized persona for target tasks, an evaluation criteria section, a guideline section, and an input section. The evaluation criteria are as simple as `\textit{the more hallucinated the content is, the higher score should be given; the more plausible the output is, the higher score should be given}'. The guideline section consists of one demonstration for each hallucination pattern. The input section comprises the model input, an input text, a set of hallucinated candidates, and instructions on the output format.

\subsection{Hallucination Pattern Guidance}\label{sec: hallucination_pattern_guidance}

Hallucination patterns depend on the domains, tasks, contexts, and questions asked.
%For example, LLMs mainly hallucinate by generating non-factual information in question answering tasks, whereas LLMs tend to hallucinate by assuming parameters in tool orchestration agents.
As such, it is critical to curate the synthetic hallucination datasets in a controlled manner such that the hallucination patterns align with model behaviors in production. Our approach achieves task-specific generation by introducing a section of \textit{Hallucination Pattern Guidance} (HPG) in the generator prompt.

The HPG module needs a set of predetermined hallucination patterns. Each pattern consists of a short description and a demonstration, including an input, a non-hallucinated output, and a hallucinated output of the pattern. With the HPG section, the generator follows the instruction to generate hallucinated candidates in a controlled rather than open-ended manner. Our approach relies on human judgment to determine the hallucination patterns in their target applications. Such patterns can be generic, \textit{e.g.}, overconfidence and non-factuality, or task-specific, such as confusing between entities in response. Practitioners have the flexibility to include the most common and relevant hallucinations by simply writing out descriptions and curating demonstrations. Moreover, it is worth noting that the predefined pattern can go beyond the conventional definition of hallucination and include any undesired LLM behaviors that we want to detect. For example, in the experiment to be presented in Section \ref{sec:experiment}, we include the pattern of nonsensical responses, where the generated responses bear no meaning in the context.

%The generation-selection pipeline is sufficiently generic to support various tasks. However, 
\subsection{Language Style Alignment}\label{sec: language_feature_alignment}

LLM generations are known to be biased, lack diversity, and misaligned with human writings \citep{LLM_generation_attribute_manipulation}. As a result, a synthetic dataset created by one LLM might be sufficiently distant from human writings or the generation of other LLMs. Since our approach leverages the golden non-hallucinated outputs, any salient distinctions in language styles like length of text or tone between hallucinated output and non-hallucinated output can be exploited as shortcuts during supervised training. 
Shortcuts impact generalization by misleading the detectors to focus on superficial features rather than hallucination patterns. 
Due to the fast-paced development of LLM applications, there is a strong case for better generalization abilities of hallucination detectors. 
Therefore, it is critical to ensure the synthetic datasets resemble the language characteristics of golden, non-hallucinated text.

To this end, we propose a \textit{Language Style Alignment} (LSA) module to align the characteristics of generated text with benchmark text. LSA is achieved by a prompt section in the generator prompt, which includes a group of itemized guidelines (see Appendix~\ref{appendix:language_feature}) on the desired language style features, e.g., writing style, length, tone, etc. With the LSA, the generator follows the instructions and generates the hallucinated candidates in a controlled manner. In general, language-style-aligned hallucinated outputs should be more challenging to detect as they are more similar to non-hallucinated outputs, except for the hallucinated content.

The challenge lies in obtaining the language style features. For one, the benchmark dataset to be aligned with can be too large to manually analyze the text characteristics. Besides, it requires expertise in linguistics to properly analyze and summarize the language style features. To overcome the challenge, we propose a \textit{Language Style Discovery} algorithm that leverages LLMs to summarize and consolidate the language style features. Specifically, the benchmark dataset is first partitioned into batches of proper size and then fed into prompted LLMs to analyze language style features. A group of language style features is produced for each batch, and language style features are merged together to form a language feature set. Then, we partition the language feature set into batches and ask a LLM to consolidate the features in each batch, which results in a smaller language feature set. The procedure continues until the desired number of language style features are obtained.

\subsection{Data Mixture}
Different LLMs exhibit different bias, diversity, and misalignment issues due to the distinctions in the pre-training corpus and preference alignment.
As a result, a synthetic dataset created by one LLM might be sufficiently distant from the ones generated by other LLMs, such that detectors trained on the synthetic dataset do not generalize even if the language style alignment is in place. 
Such generalizations for hallucination detection become more and more important due to the increased restrictions on the usage of LLM-generated data. 
% 

% Such generalizations for hallucination detection become more and more important due to the increased restrictions on the usage of LLM-generated data. For example, OpenAI restricts users from developing third-party models using the data generated by ChatGPT. For applications built with ChatGPT, developers can only resort to open-sourced LLMs with a permissive license to curate synthetic datasets to train detectors, hoping the detectors will generalize. Also, the landscape of LLM development is changing rapidly. LLM users may switch to other LLMs for various reasons. Hallucination detectors with strong generalization ability reduce the dependency on specific LLMs and thus enable faster developments and experiments.
To this end, we experiment with \textit{Data Mixture}, a simple-yet-effective scaffolding strategy, to further boost the generalization and performance robustness of detectors trained on synthetic data. Specifically, we run the generation-selection pipeline with multiple LLM generators and mix the resulting synthetic dataset to increase the training corpus diversity and mitigate bias. Note that the data mixture is independent of the pipeline design since it is applied to the resulting synthetic datasets.

\section{Experiments}\label{sec:experiment}

\subsection{Setup}

% \vspace{-5pt}
\paragraph{Benchmarks.} 

We consider three task-oriented conversational benchmarks to conduct the empirical experiments: \textbf{OpenDialKG} \cite{opendialkg}, \textbf{ReDial} \cite{redial} and \textbf{SalesBot} \cite{salesbot}. For each benchmark, we randomly sample 1000 data points from the dataset as the golden non-hallucinated samples and apply the data generation pipeline to selected samples to curate synthetic datasets. 

% \vspace{-5pt}
\paragraph{Setup. }

We run experiments with six LLMs from three model families available in AWS Bedrock, including Claude3-Sonnet, Claude3-Haiku \citep{anthropic2023modelcard}, llama2-13B, llama2-70B \citep{llama2}, Mixtral-8$\times$7B Instruct, and Mixtral-Large \citep{jiang2024mixtral}. We use the same LLM for generation and selection and refer to the resulting dataset under the corresponding LLM's name. We use Claude3-Sonnet to analyze the datasets and discover a set of language style features (see Appendix \ref{appendix:language_feature}).

We manually curate three hallucination patterns for our experiments, including non-sensical response, inconsistent entity, and irrelevant content. Each hallucination pattern is associated with a demonstration example.\footnote{We also run the experiments using five automatically curated hallucination patterns. The generalization evaluation is deferred to Table \ref{tab:auto_robustness_full} and \ref{tab:auto_cross_task_generalization_full} in Appendix \ref{appendix: experiment_result}.} Details on the hallucination patterns are in Appendix \ref{appendix:HPG}.
We generate three hallucination candidates per sample for each pattern. As a result, each synthetic dataset contains 4000 samples, including 1000 non-hallucinated responses and 3000 hallucinated responses. Detailed parameters for the generation pipeline and fine-tuning are deferred to Appendix \ref{appendix: experiment_setup}.

For the experiment with the data mixture, we evaluate two mixture strategies based on the generator portfolios: model family mixture and model size mixture. The former strategy combines synthetic datasets generated by LLMs in the same model family (Claude3, Llama2, Mixtral), and the latter combines the datasets generated by the larger models in each family (Large Combo: a mixture of Claude3-Sonnet, Llama2-70B, and Mixtral-Large) and smaller models in each family (Small Combo: a mixture of Claude2-Haiku, Llama2-13B, and Mixtral-8$\times$7B). We mix synthetic datasets through random sampling while controlling the dataset size for the sake of fair comparison. 

%\input{tables/performance_by_category_v2}
% \vspace{-5pt}
\paragraph{Evaluation.} 

\input{table_external_submission/good_hallucination_corpus_distance}

Considering the objective of hallucination detection, we evaluate our approach through two branches of metrics. Firstly, we quantify the detector performance with standard metrics for binary classification, such as the F1 score. We run the supervised detectors on the test datasets generated by the same LLMs as the training dataset (\emph{in-generator}), and the results reveal whether the hallucination is detectable and how good the detectors are.

We evaluate the generalization abilities through three pillars: 1) Cross-generator generalization is measured by the performance on test dataset generated other LLMs (\emph{out-of-generator}); 2) cross-pattern generalization is assessed on the unseen patterns (\emph{out-of-pattern}); 3) cross-task generalization is investigated by training supervised detectors on one benchmark task while evaluating on other benchmarks (\emph{out-of-task}). For performance robustness, we adopt the standard deviation of the metrics recorded on out-of-generator datasets (out-of-generator std).

% Secondly, we also compute the difference between the performance on in-generator dataset(s) and \emph{out-of-generator} datasets, i.e., datasets generated by LLMs other than the one generating the training dataset. It quantifies a detector's generalization ability across generators. Besides, we also evaluate the detector's ability to generalize on unseen hallucination patterns by using two hallucination patterns to train detectors (\emph{in-pattern}) and using one as a holdout pattern dataset (\emph{out-of-pattern}). Furthermore, to further investigate the performance generalization ability across tasks, we also evaluate their performance under the scenario where the training benchmark task is different from the testing benchmark task (\emph{out-of-task}). 

\vspace{-5pt}
\subsection{Synthetic Dataset Analysis}

% Language style alignment is designed to align the generation style with non-hallucinated samples. 
% We first quantitatively examine the efficacy of LSA by gauging the distance between synthetic hallucinated responses and non-hallucinated responses. 
We first compare the generation quality of our approach with similar baselines by gauging the distance between synthetic hallucinated and non-hallucinated responses. Datasets reporting smaller distances are considered to be of higher quality since hallucinated samples better resemble the good responses.
We utilize three metrics to quantify the distance between two corpora: Fréchet Inception Distance (FID) \citep{fid}, Zipf \citep{zipf}, and Medoid \citep{measuring_the_measuring_tool_2022}. FID quantifies the corpus distance through the Wasserstein distance between densities by fitting a continuous multivariate Gaussian to the SentenceBERT text embeddings of corpora. Zipf gauges the distance using the absolute difference between two Zipfian coefficients fitted on two corpora. Lastly, Medoid quantifies the cosine distance between corpora centroids. Besides, we compare our approach with SimPrompt \citep{YuZZMRKSZ23} where LSA is removed, and HaluEval \citep{HaluEval_LiCZNW23}.

Distances between hallucinated and non-hallucinated responses are reported in Table \ref{tab: good_hallucination_distance}. The distances are consistently smaller with our approach, \textbf{demonstrating that our approach generates hallucinated samples closer to the real human non-hallucinated samples}. Specifically, compared with SimPrompt, our hallucinated responses are 12.0\%, 11.5\% and 6.8\% closer to the good ones on average for OpenDialKG, ReDial, and SalesBot, respectively. The distance improvement is even larger when compared with HaluEval.
% Besides, we also observe that the distances are not equivalent for the three benchmarks, with the responses being most distant for SalesBot and least distant for ReDial.
The reduced corpus distance implies that our approach guides the generation to resemble the language features of non-hallucinated samples. This resemblance makes it more difficult for a detector to focus on trivial language features irrelevant to detecting hallucinations.%and thus improves the generation quality.

\subsection{Hallucination Detection Performance}

Table \ref{tab:performance_by_category} reports the average of F1 scores recorded by ICL detectors and supervised detectors (Vanilla and Mixture). For more detailed results by individual models, refer to Table~\ref{tab:performance_by_category_full} in the appendix. For ICL detectors, the LLM is assessed on the synthetic dataset generated by the same LLM. For supervised detectors, the reported performance is in-generator performance—the detectors are assessed on the test dataset generated by the same LLM as the training dataset. We find that ICL detectors still face significant challenges identifying hallucinations generated by themselves; the average F1 score is only 0.613 across the board. Fine-tuned detectors, in contrast, exhibit stronger performance consistently. For vanilla fine-tuned detectors trained on synthetic datasets generated by specific LLMs, the average F1 scores for six models are 0.920, 0.932, and 0.963 on OpenDialKG, ReDial, and SalesBot, respectively. %and the number is 0.912 for supervised detectors trained with data mixture.

Furthermore, rows of Vanilla/Mixture OP report the F1 score when the data of a hallucination column is removed from the training data — a setup to evaluate the out-of-pattern generalization. We find that supervised detectors perform worse on unseen patterns on average, but the gap is small—supervised detectors still outperform ICL detectors by a large margin. Such results mitigate the concern of misrepresenting run-time hallucinations, assuring superior performance on underrepresented patterns. 

Supervised detectors with data mixture perform slightly worse than vanilla supervised detectors on average (0.938 versus 0.912). The same pattern is observed in each hallucination pattern category and benchmark task. Since we control the sample size, the degraded performance suggests that synthetic datasets generated by different LLMs are distant in distribution, so merging datasets without scaling the sample size is at the cost of model performance. Interestingly, the conclusion holds even for datasets generated by the LLMs in the same family (see Table \ref{tab:performance_by_category_full} in Appendix \ref{appendix: experiment_result}). We conjecture that model size plays a role in generation distributions, as models in the same family usually share the training corpus.
\input{table_external_submission/performance_by_category}

% Moreover, we also find that the cross-pattern generalization ability diverges as well—detectors trained on datasets generated by more powerful models (larger models) generalize better in general, in line with the observation on out-of-generator performance. Detailed results and discussion are deferred to Table \ref{tab:cross_pattern_generalization_full} in Appendix \ref{appendix: experiment_result}.

% To further test the generalization abilities of creating detectors using our approach, we further test the detection capability on a hallucination pattern on which the detector was not trained (out-of-pattern).
% Average F1 results on unseen hallucination patterns are reported in Table \ref{tab:cross_pattern_generalization}. 
% Besides, supervised detectors generalize worst on the hallucination of entity inconsistency but generalize best on the hallucination of nonsensical responses. It suggests that the generalization performance does not solely depend on the pattern difficulty.
%While we see a much larger drop in out-of-pattern (OP) than in-pattern (IP) for the data mixture it still has a high F1 score at 85.1\%.
%We believe that data mixture OP results are true indicators of out-of-pattern generalization capabilities, as the detectors are less likely to pick up shortcuts from language styles due to the better text diversity induced by data mixture.

\subsection{Generalization Investigation}

The results in the previous section on data-mixture suggest that performance suffers because of differences in the data distribution between text generated from different LLMs, implying a generalization issue with detectors struggling with out-of-distribution text. We further explicitly investigate this in two scenarios: out-of-generator generalization and out-of-task generalization. 

\paragraph{Out-of-Generator Generalization (OGG). } 

We investigate OGG by training supervised detectors on a dataset generated by one generator (or more in the case of mixture) and testing on the datasets generated by the rest of the generators.

As shown in the performance panel of table \ref{tab:out-of-generator}, supervised detectors continue delivering superior performance than ICL detectors across the board, though slightly underperform the in-the-generator detectors.
Besides, we observe that supervised detectors trained with mixed data outperform vanilla supervised detectors on out-of-generator datasets by 0.032 on average. This indicates that data mixture is a simple yet effective approach to increasing the OGG ability. 
Moreover, mixture trained detectors outperform both SimPrompt and HaluEval by 0.011 and 0.112 respectively. 
% Moreover, both vanilla and mixture detectors outperform the baseline HaluEval. 

%For example, in the top panel, which presents the main results recorded by the supervised detectors generated from our approach with data mixture and language style alignment, the average in-generator F1 is 0.903, whereas the average out-of-generator F1 is only 0.868.

Apart from the average performance, we adopt the standard deviation of the out-of-generator F1 score to proxy the generalization robustness; a detector with robust generalization is considered more reliable to deliver consistent performance. As shown in the robustness panel, the average out-of-generator standard deviation for vanilla supervised detectors is 0.095, equivalent to 11.2\% of the out-of-generator mean. Supervised detectors with data mixture achieve a smaller out-of-generator standard deviation; the average is 0.065, equivalent to 7.4\% of the out-of-generator mean. The results confirm that the data mixture strategy improves the model's robustness when transferring supervised detectors across LLMs. Similar to the generalization performance results, mixture trained detectors achieve best-in-class generalization robustness. 
% Looking into the details, supervised detectors' generalization abilities diverge—detectors trained on Mixtral-generated datasets generalize better than detectors trained on other datasets, but the detectors trained on Llama2-generated datasets are consistent laggards. Besides, it is shown that detectors trained on datasets generated by more powerful models (larger models) generalize better in general. Since the generalization ability of supervised detectors is mainly determined by the dataset quality, we believe that the difference in generalization ability reflects an LLM's inherent generation bias and instruction-following capability. Detailed results are deferred to Table \ref{tab:robustness_full} in Appendix \ref{appendix: experiment_result}.
% \vspace{-5pt}

\input{table_external_submission/cross_model_generalization2}

% \input{table_external_submission/cross_pattern_generalization}
% \vspace{-5pt}
\paragraph{{Out-of-Task Generalization (OTG).}}

There is a practical motivation for transferring existing supervised detectors trained on one task to a new task to reduce the development burden. An ideal hallucination detector should generalize well to other tasks when the hallucination patterns are similar. We investigate this OTG ability by training supervised detectors on one benchmark task and evaluating them on others.

Performance is reported in Table \ref{tab:out-of-generator}. Vanilla trained detectors outperform the ICL detectors by 0.257 on average on the three tasks; Mixture trained detectors outperform by the margin of 0.247. Compared with in-domain scenarios, the out-of-task performance degrades by 0.066 and 0.071 using mixture and vanilla trained supervised detectors, respectively. The findings suggest that supervised detectors produced based off our data pipeline are superior alternatives to ICL in the scenarios of light-weight development, offering plug-and-play capability with great generalization. 

Besides, both vanilla and mixture trained detectors outperform the SimPrompt and HaluEval across the benchmark tasks. Vanilla trained detectors record the best performance, outperforming HaluEval and SimPrompt detectors by 0.060 and 0.043. Mixture trained detectors record a smaller lead but consistently outperform as well. We conjecture that the data mixture strategy doesn't benefit OTG because our strategy only mixes the data in the generator dimension. We believe that extending the generalization ability from one dimension (obtained by data mixture) to another presents significant potential, and we leave this exploration for future work.

% Besides, we again compare the OTG and OGG performance with HaluEval and SimPrompt. In OGG evaluation, the mixture trained model achieves the best performance and robustness, whereas the vanilla trained model falls behind the SimPrompt slightly. As for the OTG evaluation, vanilla trained detectors outperforms HaluEval and SimPrompt detectors by 0.113 and 0.043, respectively. Mixture trained detectors record a smaller lead but consistently outperforms as well. The results reinforces
% our pipeline's superiority against HaluEval and SimPrompt in terms of detector generalization. 

% \input{table_external_submission/cross_task_generalization}

\vspace{-5pt}
\subsection{Ablation Study}

The central conjecture of our method is that LSA and HPG generate non-trivial hallucinations, which are more aligned with non-hallucinated samples. The direct implication of this conjecture is \emph{it would be harder to detect hallucinations generated using LSA and HPG than without them}, as these hallucinations are more similar to non-hallucinated responses in \emph {language style}.

We test this conjecture in Table \ref{table:ablation-v2} with an ablation on LSA and HPG components. 
Note that values on the diagonal are higher as expected since the train and test sets are more similar. We observe that the average performance on test hallucinations generated with both LSA and HPG is much lower than the ones w/o LSA or w/o HPG, supporting the conjecture that the synthetic samples become easier to detect without LSA and HPG. Particularly, w/o HPG, the generated hallucinations become too trivial for the detector to detect, with an average F1 of 0.973 versus 0.908 w/ HPG across three benchmarks. LSA also makes the hallucinations harder to detect, though with a lower effect compared to HPG (0.917 average F1 w/o versus 0.908 w/ LSA).

Moreover, the detector performance also provides a lens to examine the training data quality by comparing the F1 scores in each column. Specifically, detectors trained on datasets without LSA underperform the ones with LSA (LSA + HPG) consistently across all datasets (except on test data w/o LSA) and benchmarks. Similarly, detectors trained on datasets without HPG record much lower performance compared with the ones with HPG (LSA + HPG) consistently. It suggests that synthetic data created using LSA and HPG possesses superior quality, resulting in more effective supervised detectors. These results provide strong evidence for our conjecture that LSA + HPG generate more difficult and high-quality hallucinations.

\input{table_external_submission/ablation}

\section{Related Work}

A bank of benchmarks has been curated for hallucination detection and evaluation recently. \citet{MedHaLTPalUS23} proposes a hallucination benchmark specific to LLMs in the medical domain. It consists of multiple-choice questions from various countries focusing on reasoning ability and memory ability. \citet{Factor_MuhlgayRMLRBALSS24} introduces a method for automatically creating hallucination benchmarks by perturbing factual statements. BAMBOO \citep{BAMBOO} and ScreenEval \citep{ScreenEval_LattimerC0Y23} are two benchmarks focusing on hallucination detection in the context of long texts. Rather than focusing on sentence-level hallucination detection, PHD is a benchmark designed for passage-level detection \citep{PHD_YangS023}. The most similar work to ours is HaluEval, which uses ChatGPT to create a task-specific hallucination benchmark for four tasks \cite{HaluEval_LiCZNW23}. In contrast, our work is designed to be generic to generate customized hallucination datasets for any task or domain. 
%Hallucination benchmarks designed for non-English LLMs have also emerged, such as HalluQA for Chinese LLMs \citep{halluqa_cheng2023}.

Many researches extend the naive approach to address the bias and diversity issues observed in LLM generations. \citet{YuZZMRKSZ23} demonstrates that attributed prompts (specifying attributes like length and style) outperform naive prompts in terms of the resulting model's performance. \citet{LLM_generation_attribute_manipulation} proposes Chain-of-Thoughts Attribute Manipulation (CotAM) to curate datasets from LLMs through few-shot learning. The motivation behind the approach is to create a dataset with changes only in the attribute targeted by the task. PROGEN utilizes the feedback from downstream models to guide generations via in-context examples in an iterative manner \citep{PROGEN_YeG0F0K22}. Our work extends the attribute manipulation approaches by automatically discovering the language styles by LLMs. Besides, our method degenerates to the simple baseline (SimPrompt) when LSA module is removed and the AttrPrompt when LSA is replaced with attribute guidance \citep{YuZZMRKSZ23}. 

\vspace{-2.5pt}
\section{Conclusion}

We propose a generic automated approach to generate synthetic datasets for training \emph{non-trivial} hallucination detectors. Our analysis reveals that our approach better aligns the hallucinated synthetic text with non-hallucinated benchmark samples versus existing methods~\cite{HaluEval_LiCZNW23}, in order to create non-trivial hallucination detectors.  Our experiment results show that ICL LLM detectors are only slightly above chance, and detectors trained on synthetic datasets outperform ICL LLM detectors by a large margin. Moreover, the detectors trained on synthetic datasets have cross-generator, cross-pattern, and cross-task generalization abilities, implying that our pipeline produces versatile detectors for various application scenarios.

%increase the quality of synthetic data by automatically discovering language style features in non-hallucinated benchmark samples to guide the generators.

% Our ablation study reveals that Hallucination Pattern Guidance (HPG) reduces the inherent bias by explicitly controlling the patterns of hallucination, increasing the hallucination detectors' generalization and robustness. Our method's generalization capability is shown by a high out-of-pattern and cross-task hallucination detection performance. The data mixture strategy from multiple LLMs achieves the same effect at a slight cost of in-generator performance under the same data budget constraint. Moreover, Language Style Alignment (LSA) is an effective strategy to increase the quality of synthetic data by automatically discovering language style features in non-hallucinated benchmark samples to guide the generators.

To conclude, we propose a versatile framework for curating task-specific synthetic hallucination datasets for building \textit{post-hoc} hallucination detectors. It contains an effective procedure to detect non-trivial hallucinations, using language style discovery and hallucination pattern customization to make detectors generalized and robust. We believe it paves the way for building low-effort customized hallucination detection models.



\section*{Limitation}

We rely on human judgment to curate hallucination patterns, which may not cover all the potential hallucination patterns. Consequently, the detector performance reported in our paper doesn't necessarily align with production performance. Besides, we generate an equal amount of hallucination samples for each pattern, resulting in a balanced dataset. The occurrence frequency of various patterns can, however, be significantly different.

Supervised hallucination detectors cannot solve all the types of hallucinations. Their efficacy is in question in cases where hallucination patterns require intense knowledge. Despite the fact that our pipeline can generate factuality hallucinations, the detector's performance may degenerate in such applications. 
\bibliography{synthetic_data_generation}

\appendix

\clearpage

\section{Experiment Setup}\label{appendix: experiment_setup}

\subsection{Benchmarks}
\paragraph{OpenDialKG \cite{opendialkg}.} It is a dataset of task-oriented conversations between two crowdsourcing agents engaging in a dialog about a given topic. The dataset contains conversations for two tasks: recommendation and chit-chat. The conversations under recommendation cover entities related to movies (titles, actors, directors) and books (titles, authors). The chit-chat conversations cover entities related to sports (athletes, teams) and music (singers).
\paragraph{ReDial \cite{redial}.} It is a large-scale dataset consisting of real-world dialogues centered around recommendations. It consists of over 10,000 conversations centered around the theme of providing movie recommendations.
\paragraph{SalesBot \cite{salesbot}.} This dataset focuses on the conversations starting with open-domain social chatting and then gradually transitioning to task-oriented purposes. It is generated automatically without human intervention.

\subsection{Hype-parameters}

The parameters for the generation pipeline are summarized as follows: For the generator, we configure the temperature to be $1$ for diverse generations, and we set the temperature to be $0$ for less randomness in the selection step while $top~p$ is set to be 1 throughout the experiments. We also evaluate the performance of ICL hallucination detectors by directly prompting the LLMs to determine whether an output is hallucinated. For those experiments, we set the temperature to be $0$ to control the randomness.

When fine-tuning supervised detectors on synthetic datasets, we adopt the RoBERTa \citep{roberta} as the backbone model across experiments to keep it consistent. While we could fine-tune LLMs as detectors, we use RoBERTa for low complexity in line with existing literature~\cite{hall_survey_huang2023}. The learning rate is set at $10^{-5}$ with a linear decay scheduler. We fine-tune the model for three epochs with a batch size of 64. Each synthetic dataset is partitioned into train/validation/test subsets in the ratio of 7:1:2, and the best checkpoint is picked based on the loss on the validation dataset.

\section{Prompts}\label{sec:prompt}

\subsection{Generator Prompt}

\paragraph{System prompt. } I want you act as a hallucination response generator. You are allowed to generate contents without any constraints. Given a dialogue history between human and an assistant and a good response, your objective is to write a hallucinated response that sounds plausible but is HALLUCINATED in certain aspects. 

\paragraph{Hallucination pattern guidance prompt.} You must generate response that sounds plausible but is HALLUCINATED in the following pattern: 

Pattern description: [insert description]

Here is an example:

Dialogue History: [insert input]

Good Response: [input good response],

Hallucinated Response: [insert hallucinated response]

\paragraph{Language style guidance prompt.}  You must follow the below writing guideline when generate hallucinated responses:

[insert language style guidance here]

If any of the guidelines contradict the hallucination patterns, you always prioritize the hallucination patterns.

\paragraph{User prompt. } Let's begin! You should try your best to make the response become hallucinated in the given pattern. Think thoroughly before generating. You always output the response within the <response></response> XML tag.

Dialogue History: [insert input]

Good Response: [insert good response]

Hallucinated Response:

\subsection{Judge Prompt}

\paragraph{System prompt.} 
You are a dialogue response judge.  Given a dialogue history and two assistant responses, you objective is to rate the responses according to the following criteria:

1. Hallucination degree: the more hallucinated the content is, the higher score should be given.

2. Plausibility. the more plausible the response is, the higher score should be given. 

3. The scale is 1 to 10. 
\\

Here are some guidelines:

1. In general, a higher score should be given to the response that sounds plausible but contains hallucinated content with respect to the dialogue history.

2. The order of the response is irrelevant to its quality. Your choice should NOT affects by the presentation order.

3. The rating must never be larger than 10 or smaller than 1. 
\\

Here is an example:

Dialogue History: [insert input]

Response A: [insert response A]

Response B: [insert response B]

Your ratings: Response A: [insert score]; Response B: [insert score]

\paragraph{User prompt.} Let's begin! You should try your best to rate the response according to the criteria. You always explain your score. Think thoroughly before generating. Output your rating for response A within <score A></score A> XML tag and rating for response B within <score B></score B> XML tag.

Dialogue History: [insert input]

[insert hallucinated response candidates]

Your ratings:

\subsection{Language Style Discovery Prompt}

\paragraph{Raw-data-to-feature prompt.} You are a text feature and style analyst. 

You are given a group of paired historical conversation and response. Your job is to analyze the feature and style of the response. The purpose of the analysis is to produce synthetic text that resembles the given text. 

You only analyze the response, not the historical conversation. 

You always provide a description of the observed text feature and generate explanation accordingly. 

Here are the group of historical conversation and responses:

[insert a batch of data]

Please summarize the text features and styles and give the explanation. Think throughly before outputting anything. Put the response in the following format: <feature></feature>, <explanation></explanation>

\paragraph{Feature-to-feature prompt.} 

You are a text feature and style analyst. 

You are given a group of text features summarized by different analysts for a group of historical conversation and response. You job is to consolidate, merge and refine the text features and styles. 

You always output a list of text features that summarize the group of given text features. 

Here are the group of text features, organized by analysts:

[insert a batch of language style features]

Summarize, consolidate and refine the given text features. Think thoroughly before outputting anything. Put the response in the following format: <feature></feature>, <explanation></explanation>

\section{Language Style Features}\label{appendix:language_feature}

\paragraph{OpenDialKG}
\begin{itemize}
    \item Concise and conversational responses. The responses are generally short, direct, and focused, presenting relevant information succinctly without excessive details or wordiness. They employ a friendly, informal, and conversational tone, using contractions, colloquialisms, and polite phrases like \"you're welcome\" and \"enjoy\" to create a natural, engaging dialogue. 
    \item Context awareness and relevance. The responses demonstrate an understanding of the conversational context by acknowledging and referring to the user's previous statements, addressing specific inquiries, and providing relevant recommendations, details, or follow-up questions related to the topics discussed. This context awareness ensures that the information provided is pertinent and tailored to the user's interests and needs.
    \item Factual information and domain knowledge. The responses showcase factual knowledge in specific domains, such as literature, movies, and entertainment, by providing objective details about authors, titles, genres, release dates, and related information when prompted. This factual information is presented in an objective and impartial manner, without subjective commentary or personal opinions.
    \item Conversational flow and engagement. The responses maintain a natural conversational flow by seamlessly transitioning between related topics, building upon previous statements, and engaging the user with follow-up questions or suggestions. This interactive approach helps create a cohesive and dynamic dialogue, fostering continued engagement from the user. 
    \item Limited elaboration and contextual depth. While the responses provide accurate factual information, they tend to lack deeper contextual knowledge or elaborate explanations on the topics discussed. They may convey basic details but do not delve into broader themes, significance, or complex analyses, potentially indicating limitations in the knowledge base or response generation capabilities. 
    \item  Acknowledging knowledge gaps. In cases where the assistant lacks specific information or knowledge, the responses transparently acknowledge these limitations by stating phrases like "I don't know" or "I'm afraid I don't have that information." This honesty about knowledge gaps helps build trust and credibility in the conversation.

\end{itemize}
\paragraph{ReDial}
\begin{itemize}
    \item Conversational and informal tone. The responses have a friendly, casual, and conversational tone, using contractions, colloquialisms, simple language, and a manner of expression that mimics natural human dialogue. This informal style helps create a relaxed, approachable atmosphere and builds rapport with the user.
    \item Concise and focused responses. The responses tend to be relatively concise, often consisting of just one or a few sentences. This brevity reflects a natural conversational flow, allowing a dynamic exchange while avoiding overwhelming the user with excessive information. The responses stay focused on providing relevant movie recommendations and responding directly to the user's input. 
    \item Expression of personal opinions, reactions, and anecdotes. The responses incorporate personal opinions about movies, share reactions and enthusiasm, and sometimes include anecdotes or experiences related to particular films. This personal and opinionated commentary makes the conversation feel more genuine, relatable, and engaging while fostering a sense of connection with the user. 
    \item Positive sentiment and encouraging language. The responses often use positive language, affirmative statements, and encouraging tones when recommending movies or responding to the user. This uplifting and supportive sentiment contributes to a pleasant conversational experience.
    \item Engaging the user through questions and acknowledgments. The responses engage the user by directly acknowledging their comments or questions, asking follow-up questions about preferences or opinions, and making an effort to continue the conversational flow. This engagement encourages the user's active participation and helps maintain a dynamic, interactive dialogue.
    \item Use of conversational markers and continuations. The responses employ conversational markers (e.g., "oh", "well", "yep"), transitions, and open-ended continuations to bridge ideas, maintain flow, and create a natural sense of continuity within the dialogue. 
    \item  Basic adherence to grammar and conventions. While adopting a conversational style, the responses generally follow standard rules of grammar, punctuation, and sentence structure, ensuring clarity and effective communication.
    \item Occasional humor and witty remarks. In some instances, the responses incorporate humor, witty comments, or light-hearted jokes to add entertainment value and levity to the conversation.

\end{itemize}

\paragraph{SalesBot}
\begin{itemize}
    \item Conversational flow and contextual understanding. The responses demonstrate the ability to maintain a natural conversational flow, building upon the context and details provided in the preceding dialogue. They incorporate contextual references, such as movie titles, travel details, and user preferences, to provide coherent and relevant responses tailored to the specific conversation history and user's needs.
    \item Confirmation, clarification, and follow-up questioning. The responses frequently seek confirmation and clarification from the user by rephrasing details, asking follow-up questions, or prompting for additional information. This helps ensure clear understanding and accuracy before proceeding with requested actions or providing information, while also facilitating a smooth continuation of the conversational flow.
    \item Concise and direct responses. The responses are typically concise and direct, providing relevant information or addressing the user's requests without unnecessary elaboration or fluff. They tend to be focused, with short sentences and plain vocabulary, contributing to a clear and straightforward communication style. 
    \item Providing relevant information, suggestions, and recommendations. Depending on the context, the responses may retrieve and present specific structured information, such as movie showtimes, attraction details, or reservation information. They also demonstrate the ability to provide relevant suggestions or recommendations based on the user's preferences and conversation history, offering helpful options for the user to consider.
    \item Task-oriented and procedural guidance. When specific tasks or actions are requested, such as playing media or booking reservations, the responses provide step-by-step procedural guidance to assist the user in accomplishing the desired objective. This task-oriented approach aims to be practical and helpful in achieving the user's stated goals.
    \item Polite, friendly, and personalized tone. The responses maintain a polite, friendly, and personalized tone through the use of courteous language, affirmations, and first-person pronouns. Phrases like "please", "thank you", and positive expressions contribute to a pleasant and engaging conversational experience while maintaining an appropriate level of formality for an AI assistant. 
    \item Open-ended questioning and inviting further interaction. Many responses conclude by asking open-ended questions or explicitly inviting further interaction, such as "Do you need more help?" or "Can I assist you with anything else?". This feature encourages the user to continue the conversation, make additional requests, or explore different topics, fostering an open and flexible dialogue.
\end{itemize}

\newpage
\section{Hallucination Patterns}\label{appendix:HPG}

We use three identical hallucination patterns to generate synthetic datasets for all three benchmark tasks. The details of the hallucination patterns are summarized in Table \ref{tab:hallucination_pattern_information}.

\input{table_external_submission/hallucination_patterns}

\clearpage

\section{Experiment Results}\label{appendix: experiment_result}

For the detection performance, Table~\ref{tab:performance_by_category_full} and Table~\ref{tab:cross_pattern_generalization_full} provide detailed hallucination detection performance by hallucination category, as a supplement to the performance reported in Table \ref{tab:performance_by_category}. As for the generalization investigation, detailed results on the out-of-generator generalization and robustness are reported in Table~\ref{tab:robustness_full} and results for cross-task generalization are provided in Table ~\ref{tab:cross_task_generalization_full}.

We also run the \textit{Hallucination Pattern Discovery} (HPG) algorithm to curate a list of automatic hallucination patterns. The HPG algorithm follows the design of the language style discovery algorithm but analyzes and consolidates hallucination patterns given the non-hallucinated input-output pairs. This results in 5 automatic hallucination patterns: context and instruction handling deficiencies, conversational incoherence and non-sequitur, factual hallucinations and false information, inconsistent persona and self-contradictions, lack of pragmatic understanding, and common sense. 

We run the generalization investigation based off datasets generated using the automatically curated hallucination patterns, and the results are reported in Tables \ref{tab:auto_robustness_full} and \ref{tab:auto_cross_task_generalization_full}. In line with the findings in the main, it is observed that detectors trained using our data pipeline possess better OTG and OGG generalization and robustness across the board. Such results confirm that our pipeline is agnostic to the source of hallucination patterns.

\input{table_external_submission/performance_by_category_full_appendix}

\input{table_external_submission/cross_pattern_generalization_full_appendix}

\input{table_external_submission/cross_model_generalization_full_appendix}
\input{table_external_submission/cross_task_generalization_full_appendix}

\input{table_external_submission/auto_pattern_robustness_full_appendix}

\input{table_external_submission/auto_cross_task_generalization_full_appendix}

\end{document}

%% file: table_external_submission/good_hallucination_corpus_distance.tex
\begin{table}[t]
\centering
\footnotesize

\resizebox{0.49\textwidth}{!}{
\begin{tabular}{lcccc}
\toprule
& FID ($\downarrow$) &  Mediod  ($\downarrow$)&  Zipf  ($\downarrow$)& Ave  ($\downarrow$) \\
\midrule
\multicolumn{5}{l}{\textit{OpenDialKG}}   \\
Ours & 0.420& 0.278& 0.072& 0.256 \\
SimPrompt & 0.474& 0.318& 0.081& 0.291 \\
HaluEval  & 0.602& 0.395& 0.089& 0.362 \\
\midrule
\multicolumn{5}{l}{\textit{ReDial}}   \\
Ours      & 0.333& 0.205& 0.065& 0.201 \\
SimPrompt & 0.377& 0.230& 0.075& 0.227 \\
HaluEval  & 0.395& 0.26& 0.056& 0.237 \\
\midrule
\multicolumn{5}{l}{\textit{SalesBot}}   \\
Ours      & 0.622& 0.299& 0.271& 0.398 \\
SimPrompt & 0.672& 0.326& 0.283& 0.427 \\
HaluEval &  0.659& 0.356& 0.238& 0.418 \\
\bottomrule
\end{tabular}
}
\vspace{-6pt}
\caption{{\small Average corpus distance between non-hallucinated samples and hallucinated generations over six generators. Lower the distance the better a method is ($\downarrow$).
}}
\vspace{-15pt}
\label{tab: good_hallucination_distance}
\end{table}

%% file: table_external_submission/performance_by_category.tex
\begin{table}[t]
\centering

\resizebox{0.49\textwidth}{!}{\begin{tabular}{lcccc}
\toprule
& {Entity Incon.} & {Non. Resp.} & {Irre. Cont.} & {Overall} \\
\midrule
\multicolumn{5}{l}{\textit{OpenDialKG}} \\ 
ICL          & 0.670             & 0.587              & 0.551             & 0.638 \\
Vanilla      & \textbf{0.858}    & \textbf{0.932}     & \underline{0.947} & \textbf{0.920} \\
Vanilla OP   & 0.451             & \textbf{0.932}     & 0.931             & 0.771 \\
Mixture      & \underline{0.829} & \underline{0.931}  & \textbf{0.948}    & \underline{0.908} \\

Mixture OP   & 0.762             & 0.865              & 0.896             & 0.841 \\
\midrule
\multicolumn{5}{l}{\textit{ReDial}} \\ 
ICL          & 0.615            & 0.581             & 0.532             & 0.606 \\
Vanilla      & \textbf{0.849}   & \textbf{0.969}    & \textbf{0.975}    & \textbf{0.932} \\
Vanilla OP   & 0.719            & \underline{0.965} & 0.951             & 0.879 \\
Mixture      & \underline{0.793}& 0.957             & \underline{0.966} & \underline{0.913} \\

Mixture OP & 0.773              & 0.917             & 0.94              & 0.877 \\
\midrule
\multicolumn{5}{l}{\textit{SalesBot}} \\ 
ICL          & 0.616            & 0.559             & 0.505             & 0.595 \\
Vanilla      & \textbf{0.932}   & \textbf{0.978}    & \textbf{0.987}    & \textbf{0.963}  \\
 Vanilla OP  & 0.751            & \underline{0.975} & 0.977             & 0.901 \\
Mixture      & \underline{0.904}& 0.973             & \underline{0.984} & \underline{0.950} \\

Mixture OP   & 0.822            & 0.935 & 0.949 & 0.902 \\
\bottomrule
\end{tabular}
}
\vspace{-6pt}
\caption{{\small Hallucination detection performance by category. Reported are average F1 scores over six generators and five data mixture strategies. OP indicates that the data of the column pattern is removed from the training data. \textbf{Bold} indicates the best in class, and \underline{underline} signals the second. 
}}
\vspace{-15pt}
\label{tab:performance_by_category}
\end{table}

%% file: table_external_submission/cross_model_generalization2.tex
\begin{table}[t]
\centering
\footnotesize
\resizebox{0.49\textwidth}{!}{\begin{tabular}{lcccc}
\toprule
       & OpenDialKG & ReDial & SalesBot & Overall\\
\midrule
& \multicolumn{4}{c}{\textbf{Out-of-Generator Generalization}} \\
\midrule
% \midrule
\multicolumn{3}{l}{\textit{Performance} ($\uparrow$)} \\
Vanilla          & {0.813}  &  0.830             &  {0.895}  & {0.846} \\
Mixture          & \textbf{0.859}      &  \textbf{0.869}    &  \textbf{0.907}   & \textbf{0.878} \\
SimPrompt        & \underline{0.839}   &  \underline{0.856} &   \underline{0.908} & \underline{0.867}\\
HaluEval         & 0.659              &  {0.839} &  0.783            & 0.760 \\
\midrule
\multicolumn{3}{l}{\textit{Robustness} ($\downarrow$)} \\
Vanilla          & 0.122             & 0.090               & 0.073 & 0.095 \\
Mixture          & \underline{0.079} & \textbf{0.063}      & \textbf{0.054} & \textbf{0.065} \\
SimPrompt        & 0.111             & \underline{0.078}   & \underline{0.064} & \underline{0.084} \\
HaluEval.        & \textbf{0.015}    & 0.193               & 0.143  & 0.117\\
\midrule
& \multicolumn{4}{c}{\textbf{Out-of-Task Generalization}} \\
\midrule
\multicolumn{3}{l}{\textit{Performance} ($\uparrow$)} \\
Vanilla & \underline{0.872} & \textbf{0.859} & \textbf{0.871} & \textbf{0.867} \\
Mixture & {0.869} & \underline{0.854} & \underline{0.850} & \underline{0.858} \\
SimPrompt & 0.842 & 0.813 & 0.818 & 0.824 \\
HaluEval & \textbf{0.912} & 0.745 & 0.766 & 0.807\\
\bottomrule
\end{tabular}
}
\vspace{-6pt}
\caption{{\small Out-of-generator \& out-of-task generalization. The reported metrics in the performance panel are F1 scores on out-of-generator and out-of-task datasets, and the metric for robustness is the standard deviation of F1 scores on out-of-generator datasets. \textbf{Bold} indicates the best in class, and \underline{underline} signals the second in each section. 
}}
\vspace{-15pt}
\label{tab:out-of-generator}
\end{table}

%% file: table_external_submission/ablation.tex
\begin{table}[h]
\centering
\footnotesize
\resizebox{0.49\textwidth}{!}{
\begin{tabular}{ccccc}
\toprule
&  \multicolumn{4}{c}{Train Set}   \\
Test Set  & LSA + HPG & w/o LSA & w/o HPG & Ave \\
\midrule
\multicolumn{4}{l}{\textit{OpenDialKG}}   \\
 % \midrule
 LSA + HPG   &  0.920       &  0.911  & 0.852   & 0.894   \\
 w/o LSA     &  0.914       & 0.945   & 0.845   & 0.901    \\
 w/o HPG     &  0.946       &  0.944  &  0.976  & 0.955   \\
\midrule
\multicolumn{4}{l}{\textit{ReDial}}   \\
% \midrule
% &  \multicolumn{4}{c}{Train Set}   \\
% Test Set  & LSA + HPG & w/o LSA & w/o HPG & Ave \\
 % \midrule
 LSA + HPG   &    0.932     & 0.922     &  0.839  &  0.898  \\
 w/o LSA     &    0.932     &  0.945  & 0.822   & 0.900   \\
 w/o HPG     &    0.982     & 0.973   &  0.993  &  0.983  \\
\midrule
\multicolumn{4}{l}{\textit{SalesBot}}   \\
% \midrule
% &  \multicolumn{4}{c}{Train Set}   \\
% Test Set  & LSA + HPG & w/o LSA & w/o HPG & Ave \\
 % \midrule
 LSA + HPG   & 0.963    & 0.956  & 0.878  & 0.932   \\
 w/o LSA     & 0.975    & 0.977   & 0.898  & 0.950    \\
 w/o HPG     & 0.978    & 0.975 & 0.988  &  0.980 \\
\midrule
\end{tabular}}
\vspace{-10pt}
\caption{\small Ablation study. 
Column names denote the generation setup for train datasets and row names denote the setup for test datasets.
% This table reports the detector performance trained and evaluated on datasets generated with different modules. Column names denote the pipeline setup used to generate training datasets, and row names represent the pipeline setup for test datasets. 
}
\vspace{-15pt}
\label{table:ablation-v2}
\end{table}

%% file: table_external_submission/hallucination_patterns.tex
%\newpage
\begin{table*}[h]
    \centering
    \tiny
    
    \begin{tabular*}{0.98\textwidth}{@{\extracolsep{\fill}}p{0.1\textwidth} | p{0.85\textwidth}}
\toprule
Pattern & Entity Inconsistency \\
\hline
Description & The entity in the response is not consistent with the dialogue history.\\
Prompt & You write a response to human but you replace the true entity with a dissimilar entity. \\
Input & user: Do you know Calvin Harris? assistant: Yes he is a composer/DJ.  Some of his work is  Where Have You Been and Yeah x3.  Do you know his work? user: No. I don't know him. Do you like any of his work?. I can give a try. assistant: His record label is Ultra Music.  I don't know a lot of his work but am curious.  Are you going to listen to some of his stuff? user: Yes. I am thinking of listening some of his works.  Are you going to listen any of his songs? assistant: I am going to try Yeah 3x user: Do you know the meaning of 3x in Yeah 3x? Just wondering. \\
Good response & assistant: Chris Brown sings it and it was released in 2010, not sure of the meaning. \\
Hallucinated response & assistant: LeBron James sings it and it was released in 2010, not sure of the meaning. \\
\midrule
Pattern & Irrelevant Content \\
\hline
Description & The response contains irrelevant content to the dialogue history. \\
Prompt & You write a response that is disconnected with the context of the dialogue history. \\
Input & user: Do you know Calvin Harris? assistant: Yes he is a composer/DJ.  Some of his work is  Where Have You Been and Yeah x3.  Do you know his work? user: No. I don't know him. Do you like any of his work?. I can give a try. assistant: His record label is Ultra Music.  I don't know a lot of his work but am curious.  Are you going to listen to some of his stuff? user: Yes. I am thinking of listening some of his works.  Are you going to listen any of his songs? assistant: I am going to try Yeah 3x user: Do you know the meaning of 3x in Yeah 3x? Just wondering. \\
Good response & assistant: Chris Brown sings it and it was released in 2010, not sure of the meaning. \\
Hallucinated response & assistant: Yeah, 3x means three times or thrice in mathematics.\\
\midrule
Pattern & Nonsensical Response \\
\hline
Description & The response bears no meanings or useful information given the context of the dialogue history.\\
Prompt & You write a response that is nonsensical to the dialogue history and disrupts the dialogue flow. \\
Input & user: Do you know Calvin Harris? assistant: Yes he is a composer/DJ.  Some of his work is  Where Have You Been and Yeah x3.  Do you know his work? user: No. I don't know him. Do you like any of his work?. I can give a try. assistant: His record label is Ultra Music.  I don't know a lot of his work but am curious.  Are you going to listen to some of his stuff? user: Yes. I am thinking of listening some of his works.  Are you going to listen any of his songs? assistant: I am going to try Yeah 3x user: Do you know the meaning of 3x in Yeah 3x? Just wondering. \\
Good response & assistant: Chris Brown sings it and it was released in 2010, not sure of the meaning. \\
Hallucinated response & assistant: Chris Brown sings it and it was released in 2010.\\
\bottomrule
    \end{tabular*}
    \caption{Hallucination pattern guidance information. }
    \label{tab:hallucination_pattern_information}
\end{table*}

%% file: table_external_submission/performance_by_category_full_appendix.tex
\begin{table*}[b!]
\centering
\tiny

\resizebox{0.95\textwidth}{!}{
\begin{tabular}{lcccc|cccc|cccc}
\toprule
\multicolumn{5}{c|}{\textbf{OpenDialKG}} & \multicolumn{4}{c|}{\textbf{ReDial}} & \multicolumn{4}{c}{\textbf{SalesBot}} \\
\midrule
& Enty Incn & Non. Resp. & Irre. Cont. & Overall & Enty Incn & Non. Resp. & Irre. Cont. & Overall & Enty Incn & Non. Resp. & Irre. Cont. & Overall \\
\midrule
\multicolumn{5}{l}{\textit{In-context-learning detectors}} \\
Llama2-13B & 0.37& 0.365& 0.359& 0.401 & 0.332& 0.276& 0.32& 0.351 & 0.285& 0.218& 0.252& 0.302 \\
Llama2-70B & 0.786& 0.8& 0.8& 0.828 & 0.687& 0.664& 0.681& 0.709 & 0.573& 0.558& 0.503& 0.562 \\
Mixtral-87B & 0.478& 0.11& 0.133& 0.307 & 0.335& 0.133& 0.087& 0.252 & 0.571& 0.394& 0.222& 0.448 \\
Mixtral-Large & 0.854& 0.831& 0.807& 0.837 & 0.809& 0.869& 0.833& 0.833 & 0.814& 0.82& 0.824& 0.835 \\
Claude3-Haiku & 0.789& 0.792& 0.692& 0.792 & 0.833& 0.833& 0.724& 0.816 & 0.743& 0.696& 0.66& 0.737 \\
Claude3-Sonnet & 0.741& 0.627& 0.517& 0.662  & 0.694& 0.71& 0.549& 0.675 & 0.712& 0.667& 0.569& 0.684 \\
\midrule
\multicolumn{5}{l}{\textit{Vanilla supervised detectors}} \\
Llama2-13B & 0.982 & 0.985 & 0.975 & 0.977 & 0.977 & 0.985 & 0.99 & 0.982 & 0.995 & 0.993 & 0.998 & 0.995 \\
Llama2-70B & 0.878 & 0.938 & 0.945 & 0.932 & 0.898 & 0.945 & 0.953 & 0.925 & 0.94 & 0.975 & 0.99 & 0.962 \\
Mixtral-87B & 0.797 & 0.866 & 0.922 & 0.870 & 0.951 & 0.988 & 0.995 & 0.973 & 0.956 & 0.985 & 0.995 & 0.972 \\
Mixtral-Large & 0.819 & 0.893 & 0.927 & 0.891 & 0.856 & 0.948 & 0.95 & 0.93 & 0.921 & 0.961 & 0.968 & 0.951 \\
Claude3-Haiku & 0.907 & 0.971 & 0.971 & 0.955 & 0.878 & 0.99 & 0.99 & 0.946 & 0.943 & 0.99 & 0.99 & 0.973 \\
Claude3-Sonnet & 0.764 & 0.941 & 0.941 & 0.895 & 0.533 & 0.956 & 0.973 & 0.834 & 0.838 & 0.968 & 0.983 & 0.923 \\
% \midrule
% Average & 0.858 & 0.932 & 0.947 & 0.920 \\
\midrule
\multicolumn{5}{l}{\textit{Supervised detectors with data mixture}}  \\
Claude 3 & 0.763 & 0.958 & 0.959 & 0.904 & 0.606 & 0.961 & 0.973 & 0.865 & 0.818 & 0.97 & 0.975 & 0.918 \\ 
Mixtral & 0.776 & 0.858 & 0.904 & 0.862 & 0.824 & 0.957 & 0.969 & 0.924  & 0.918 & 0.966 & 0.98 & 0.951 \\ 
Llama 2 & 0.924 & 0.971 & 0.97 & 0.95 & 0.926 & 0.96 & 0.964 & 0.949  & 0.965 & 0.979 & 0.99 & 0.976 \\
Small Combo & 0.877 & 0.948 & 0.964 & 0.929 & 0.908 & 0.973 & 0.973 & 0.958& 0.933 & 0.989 & 0.993 & 0.966 \\ 
Large Combo & 0.806 & 0.922 & 0.946 & 0.893 & 0.699 & 0.935 & 0.953 & 0.873 & 0.886 & 0.962 & 0.98 & 0.938 \\ 
\bottomrule
\end{tabular}
}
\caption{{\small Detailed results for hallucination detection performance by category. Reported are F1 scores. 
}}
\label{tab:performance_by_category_full}
\end{table*}

%% file: table_external_submission/cross_pattern_generalization_full_appendix.tex
\begin{table*}[b!]
\centering
\tiny
\resizebox{0.95\textwidth}{!}{
\begin{tabular}{lcccc|cccc|cccc}
\toprule
\multicolumn{5}{c|}{\textbf{OpenDialKG}} & \multicolumn{4}{c|}{\textbf{ReDial}} & \multicolumn{4}{c}{\textbf{SalesBot}} \\
\midrule
& Enty Incn & Non. Resp. & Irre. Cont. & Overall & Enty Incn & Non. Resp. & Irre. Cont. & Overall & Enty Incn & Non. Resp. & Irre. Cont. & Overall \\
\midrule
\multicolumn{12}{l}{\textit{Vanilla supervised detectors}} \\
Llama2-13B & 0.261  & 0.93  & 0.97  & 0.72 & 0.595  & 0.978  & 0.993  & 0.855 & 0.619  & 0.987  & 0.995  & 0.867 \\
Llama2-70B& 0.495  & 0.95  & 0.943  & 0.796 & 0.751  & 0.971  & 0.955  & 0.892 & 0.771  & 0.966  & 0.98  & 0.906 \\
Mixtral-87B&  0.594  & 0.93  & 0.921  & 0.815 & 0.729  & 0.978  & 0.976  & 0.895 & 0.758  & 0.995  & 0.978  & 0.911 \\
Mixtral-Large& 0.709  & 0.893  & 0.858  & 0.82 & 0.85  & 0.942  & 0.923  & 0.905 & 0.911  & 0.955  & 0.962  & 0.942 \\
Claude3-Haiku& 0.181  & 0.957  & 0.957  & 0.698 & 0.66  & 0.971  & 0.935  & 0.856 & 0.664  & 0.983  & 0.976  & 0.874 \\
Claude3-Sonnet& 0.465  & 0.932  & 0.935  & 0.777 & 0.731  & 0.951  & 0.925  & 0.869 & 0.785  & 0.964  & 0.973  & 0.907 \\
\midrule
% Average & 0.451 & 0.932 & 0.931 & 0.771 & 0.719 & 0.965 & 0.951 & 0.879 & 0.751 & 0.975 & 0.977 & 0.901 \\
% \midrule 
\multicolumn{12}{l}{\textit{Supervised detectors with data mixture}}\\
Claude3 & 0.705  & 0.857  & 0.886 & 0.816 & 0.729  & 0.914  & 0.961 & 0.868 & 0.808  & 0.92  & 0.947 & 0.892 \\
Mixtral & 0.795  & 0.837  & 0.867 & 0.833 & 0.804  & 0.899  & 0.941 & 0.882 & 0.873  & 0.948  & 0.955 & 0.926 \\
Llama2 & 0.733  & 0.869  & 0.913 & 0.838 & 0.76  & 0.924  & 0.937 & 0.874 & 0.767  & 0.931  & 0.95 & 0.883 \\
Small Combo & 0.775  & 0.875  & 0.925 & 0.858 & 0.755  & 0.917  & 0.931 & 0.868 & 0.8  & 0.95  & 0.967 & 0.906 \\
Large Combo & 0.8  & 0.887  & 0.888 & 0.858 & 0.816  & 0.93  & 0.93 & 0.892 & 0.862  & 0.925  & 0.924 & 0.904 \\
% Average & 0.762 & 0.865 & 0.896  & 0.841 & 0.773 & 0.917 & 0.94 & 0.877 & 0.822 & 0.935 & 0.949 & 0.902 \\
\bottomrule
\end{tabular}
}
\caption{{\small Detailed results for out-of-pattern performance using manual hallucination patterns. Reported are F1 scores.
}}
\label{tab:cross_pattern_generalization_full}
\end{table*}

%% file: table_external_submission/cross_model_generalization_full_appendix.tex
\begin{table*}[b!]
\centering
\tiny

\resizebox{0.95\textwidth}{!}{
\begin{tabular}{lcccc|cccc|cccc}
\toprule
\multicolumn{5}{c|}{\textbf{OpenDialKG}} & \multicolumn{4}{c|}{\textbf{ReDial}} & \multicolumn{4}{c}{\textbf{SalesBot}} \\
\midrule
& IG (Mean) & OG Mean & $\Delta$ &  OG Std & IG (Mean) & OG Mean & $\Delta$ &  OG Std & IG (Mean) & OG Mean & $\Delta$ &  OG Std \\
\midrule
\multicolumn{5}{l}{\textit{Vanilla supervised detectors}} \\
Llama2-13B & 0.977 & 0.651 & -0.326 & 0.245 &  0.982 & 0.666 & -0.315 & 0.153 & 0.995 & 0.806 & -0.189 & 0.106 \\
Llama2-70B & 0.932 & 0.814 & -0.118 & 0.14 & 0.925 & 0.856 & -0.069 & 0.084 & 0.962 & 0.892 & -0.07 & 0.067 \\
Mixtral-87B & 0.87 & 0.841 & -0.029 & 0.107 & 0.973 & 0.796 & -0.177 & 0.118 & 0.972 & 0.895 & -0.076 & 0.078 \\
Mixtral-Large & 0.891 & 0.881 & -0.011 & 0.037 & 0.93 & 0.915 & -0.015 & 0.036  & 0.951 & 0.917 & -0.034 & 0.085 \\
Claude3-Haiku & 0.955 & 0.797 & -0.158 & 0.147 & 0.946 & 0.838 & -0.108 & 0.100 & 0.973 & 0.896 & -0.077 & 0.081 \\
Claude3-Sonnet & 0.895 & 0.896 & 0.001 & 0.053 & 0.834 & 0.907 & 0.073 & 0.047 & 0.923 & 0.96 & 0.037 & 0.024 \\
\midrule
\multicolumn{5}{l}{\textit{Supervised detectors with data mixture}} \\
Claude3 & 0.904 & 0.847 & -0.057 & 0.096 & 0.865 & 0.864 & -0.001 & 0.099 & 0.918 & 0.898 & -0.02 & 0.08 \\
Mixtral & 0.862 & 0.909 & 0.047 & 0.027 & 0.924 & 0.912 & -0.012 & 0.06 & 0.951 & 0.96 & 0.009 & 0.03 \\
Llama2 & 0.95 & 0.794 & -0.156 & 0.136 & 0.949 & 0.82 & -0.128 & 0.071 & 0.976 & 0.843 & -0.133 & 0.058 \\
Small Combo & 0.929 & 0.826 & -0.103 & 0.098 & 0.958 & 0.815 & -0.142 & 0.06 & 0.966 & 0.882 & -0.085 & 0.064 \\
Large Combo & 0.893 & 0.921 & 0.028 & 0.039 & 0.873 & 0.935 & 0.062 & 0.023  & 0.938 & 0.955 & 0.017 & 0.037 \\
\bottomrule
\end{tabular}
}
\caption{{\small Detailed results for out-of-generator generalization using manual hallucination patterns. Reported are F1 scores. IG stands for in-the-Generator performance while OG stands for out-of-generator.  
}}
\label{tab:robustness_full}
\end{table*}

%% file: table_external_submission/cross_task_generalization_full_appendix.tex
\newcolumntype{L}{>{\centering\arraybackslash}m{3cm}}
\begin{table*}[h]
\centering
\footnotesize
\resizebox{0.8\textwidth}{!}{
\begin{tabular}{lLLLL}
\toprule
& Entity Inconsistency & Nonsensical Response & Irrelevant Content & Overall\\
\midrule
\multicolumn{5}{c}{\textbf{Vanilla supervised detectors}} \\
\midrule
\textit{OpenDialKG} $\to$ \textit{ReDial} & 0.799 & 0.883 & 0.885 & 0.887 \\
\textit{OpenDialKG} $\to$ \textit{SalesBot} & 0.771 & 0.886 & 0.911 & 0.863 \\
\textit{ReDial} $\to$ \textit{OpenDialKG} & 0.758 & 0.919 & 0.949 & 0.869 \\
\textit{ReDial} $\to$ \textit{SalesBot} & 0.771 & 0.941 & 0.953 & 0.878 \\
\textit{SalesBot} $\to$ \textit{Redial} & 0.764 & 0.782 & 0.783 & 0.83 \\
\textit{SalesBot} $\to$ \textit{OpenDialKG} & 0.811 & 0.829 & 0.833 & 0.874 \\
\midrule
\multicolumn{5}{c}{\textbf{Supervised detectors with data mixture}} \\
\midrule
\textit{OpenDialKG} $\to$ \textit{ReDial} & 0.783 & 0.898 & 0.903 & 0.884 \\
\textit{OpenDialKG} $\to$ \textit{SalesBot} & 0.697 & 0.853 & 0.881 & 0.816 \\
\textit{ReDial} $\to$ \textit{OpenDialKG} & 0.767 & 0.916 & 0.94 & 0.876 \\
\textit{ReDial} $\to$ \textit{SalesBot} & 0.743 & 0.939 & 0.97 & 0.883 \\
\textit{SalesBot} $\to$ \textit{Redial} & 0.758 & 0.775 & 0.776 & 0.823 \\
\textit{SalesBot} $\to$ \textit{OpenDialKG} & 0.794 & 0.815 & 0.821 & 0.862 \\
\midrule
\multicolumn{5}{c}{\textbf{SimPrompt}} \\
\midrule
\textit{OpenDialKG} $\to$ \textit{ReDial} & 0.842 & 0.908 & 0.924 & 0.909 \\
\textit{OpenDialKG} $\to$ \textit{SalesBot} & 0.8 & 0.901 & 0.922 & 0.873 \\
\textit{ReDial} $\to$ \textit{OpenDialKG} & 0.860 & 0.95 & 0.959 & 0.921 \\
\textit{ReDial} $\to$ \textit{SalesBot} & 0.836 & 0.939 & 0.965 & 0.905 \\
\textit{SalesBot} $\to$ \textit{Redial} & 0.783 & 0.793 & 0.793 & 0.845 \\
\textit{SalesBot} $\to$ \textit{OpenDialKG} & 0.824 & 0.83 & 0.832 & 0.880 \\
\midrule
\multicolumn{5}{c}{\textbf{HaluEval}} \\
\midrule
\textit{OpenDialKG} $\to$ \textit{ReDial} & - & - & - & 0.641 \\
\textit{OpenDialKG} $\to$ \textit{SalesBot}  & - & - & - & 0.642 \\
\textit{ReDial} $\to$ \textit{OpenDialKG} & - & - & - & 0.913 \\
\textit{ReDial} $\to$ \textit{SalesBot} & - & - & - & 0.888\\
\textit{SalesBot} $\to$ \textit{Redial}& - & - & - & 0.852\\
\textit{SalesBot} $\to$ \textit{OpenDialKG} & - & - & - & 0.910 \\
\bottomrule
\end{tabular}
}
% \vspace{-6pt}
\caption{{\small Results of out-of-task generalization using manually curated hallucination patterns. Reported are average F1 scores across six generators using vanilla trained detectors. }}
\label{tab:cross_task_generalization_full}
\end{table*}

%% file: table_external_submission/auto_pattern_robustness_full_appendix.tex
\begin{table*}[b!]
\centering
\tiny

\resizebox{0.95\textwidth}{!}{
\begin{tabular}{lcccc|cccc|cccc}
\toprule
\multicolumn{5}{c|}{\textbf{OpenDialKG}} & \multicolumn{4}{c|}{\textbf{ReDial}} & \multicolumn{4}{c}{\textbf{SalesBot}} \\
\midrule
& IG (Mean) & OG Mean & $\Delta$ &  OG Std & IG (Mean) & OG Mean & $\Delta$ &  OG Std & IG (Mean) & OG Mean & $\Delta$ &  OG Std \\
\midrule
\multicolumn{5}{l}{\textit{Vanilla supervised detectors}} \\
Llama2-13B     & 1.000 & 0.994 & -0.006 & 0.012 & 0.996 & 0.771 & -0.224 & 0.246 & 0.999 & 0.883 & -0.116 & 0.185 \\
Llama2-70B     & 0.999 & 0.993 & -0.006 & 0.012 & 0.994 & 0.906 & -0.088 & 0.102 & 0.998 & 0.941 & -0.058 & 0.094 \\
Mixtral-87B    & 0.987 & 0.994 & 0.008  & 0.003 & 0.987 & 0.97  & -0.017 & 0.040 & 0.983 & 0.97 & -0.013 & 0.049 \\
Mixtral-Large  & 1.000 & 0.984 & -0.016 & 0.019 & 0.972 & 0.959 & -0.013 & 0.052 & 0.966 & 0.906 & -0.06 & 0.108   \\
Claude3-Haiku  & 0.999 & 0.978 & -0.021 & 0.021 & 1.000 & 0.788 & -0.212 & 0.237 & 0.997 & 0.749 & -0.248 & 0.244 \\
Claude3-Sonnet & 1.000 & 0.987 & -0.013 & 0.018 & 0.998 & 0.899 & -0.100 & 0.112 & 1.000 & 0.922 & -0.078 & 0.124 \\
\midrule
\multicolumn{5}{l}{\textit{Supervised detectors with data mixture}} \\
Claude3      & 1.000 & 0.981 & -0.019 & 0.018 & 0.995 & 0.799 & -0.196 & 0.191 & 0.999 & 0.895 & -0.104 & 0.142 \\
Llama2       & 1.000 & 0.990 & -0.009 & 0.015 & 0.994 & 0.878 & -0.116 & 0.137 & 0.996 & 0.869 & -0.128 & 0.176 \\
Mixtral      & 0.991 & 0.999 & 0.008  & 0.001 & 0.972 & 0.99  & 0.018  & 0.003 & 0.969 & 0.990 & 0.022  & 0.002 \\
Small Combo   & 0.993 & 0.999 & 0.006  & 0.000 & 0.991 & 0.914 & -0.077 & 0.101 & 0.993 & 0.93 & -0.063 & 0.094 \\
Large Combo   & 1.000 & 0.986 & -0.014 & 0.019 & 0.982 & 0.943 & -0.039 & 0.059 & 0.983 & 0.992 & 0.009 & 0.007 \\
\bottomrule
\end{tabular}
}
\caption{{\small Detailed results for out-of-generator generalization using automatically generated hallucination patterns. Reported are F1 scores. IG stands for in-the-Generator performance while OG stands for out-of-generator.  
}}
\label{tab:auto_robustness_full}
\end{table*}

%% file: table_external_submission/auto_cross_task_generalization_full_appendix.tex
\newcolumntype{L}{>{\centering\arraybackslash}m{3cm}}
\begin{table*}[h]
\centering
\footnotesize
\resizebox{0.98\textwidth}{!}{
\begin{tabular}{lLLLLL}
\toprule
& Context and Instruction Handling Deficiencies & Conversational Incoherence and Non-Sequiturs & Factual Hallucinations and False Information & Inconsistent Persona and Self-Contradictions & Lack of Pragmatic Understanding and Common Sense \\
\midrule
\textit{OpenDialKG} $\to$ \textit{ReDial} & 0.669 & 0.669 & 0.669 & 0.669 & 0.669  \\
\textit{OpenDialKG} $\to$ \textit{SalesBot} & 0.667 & 0.667 & 0.667 & 0.667 & 0.667 \\
\textit{ReDial} $\to$ \textit{OpenDialKG} & 0.925 & 0.965 & 0.96 & 0.95 & 0.954 \\
\textit{ReDial} $\to$ \textit{SalesBot} & 0.924 & 0.977 & 0.961 & 0.947 & 0.949 \\
\textit{SalesBot} $\to$ \textit{Redial} & 0.862 & 0.862 & 0.862 & 0.861 & 0.861 \\
\textit{SalesBot} $\to$ \textit{OpenDialKG} & 0.908 & 0.925 & 0.925 & 0.922 & 0.919 \\
\bottomrule
\end{tabular}
}
% \vspace{-6pt}
\caption{{\small Results of out-of-task generalization using automatically generated hallucination patterns. Reported are average F1 scores across six generators using vanilla trained detectors.  }}
% \vspace{-15pt}
\label{tab:auto_cross_task_generalization_full}
\end{table*}

%% file: main_acl_submission.bbl
\begin{thebibliography}{31}
\expandafter\ifx\csname natexlab\endcsname\relax\def\natexlab#1{#1}\fi

\bibitem[{Anthropic(2024)}]{anthropic2023modelcard}
Anthropic. 2024.
\newblock \href
  {https://www-cdn.anthropic.com/de8ba9b01c9ab7cbabf5c33b80b7bbc618857627/Model_Card_Claude_3.pdf}
  {The claude 3 model family: Opus, sonnet, haiku}.
\newblock Technical report, Anthropic.

\bibitem[{Chiu et~al.(2022)Chiu, Li, Lin, and Chen}]{salesbot}
Ssu Chiu, Maolin Li, Yen{-}Ting Lin, and Yun{-}Nung Chen. 2022.
\newblock \href {https://doi.org/10.18653/V1/2022.ACL-LONG.425} {Salesbot:
  Transitioning from chit-chat to task-oriented dialogues}.
\newblock In \emph{Proceedings of the 60th Annual Meeting of the Association
  for Computational Linguistics (Volume 1: Long Papers), {ACL} 2022, Dublin,
  Ireland, May 22-27, 2022}, pages 6143--6158. Association for Computational
  Linguistics.

\bibitem[{Dong et~al.(2023)Dong, Tang, Li, Zhao, and Wen}]{BAMBOO}
Zican Dong, Tianyi Tang, Junyi Li, Wayne~Xin Zhao, and Ji{-}Rong Wen. 2023.
\newblock \href {https://doi.org/10.48550/ARXIV.2309.13345} {{BAMBOO:} {A}
  comprehensive benchmark for evaluating long text modeling capacities of large
  language models}.
\newblock \emph{CoRR}, abs/2309.13345.

\bibitem[{Heusel et~al.(2017)Heusel, Ramsauer, Unterthiner, Nessler, and
  Hochreiter}]{fid}
Martin Heusel, Hubert Ramsauer, Thomas Unterthiner, Bernhard Nessler, and Sepp
  Hochreiter. 2017.
\newblock \href
  {https://proceedings.neurips.cc/paper/2017/hash/8a1d694707eb0fefe65871369074926d-Abstract.html}
  {Gans trained by a two time-scale update rule converge to a local nash
  equilibrium}.
\newblock In \emph{Advances in Neural Information Processing Systems 30: Annual
  Conference on Neural Information Processing Systems 2017, December 4-9, 2017,
  Long Beach, CA, {USA}}, pages 6626--6637.

\bibitem[{Holtzman et~al.(2020)Holtzman, Buys, Du, Forbes, and Choi}]{zipf}
Ari Holtzman, Jan Buys, Li~Du, Maxwell Forbes, and Yejin Choi. 2020.
\newblock \href {https://openreview.net/forum?id=rygGQyrFvH} {The curious case
  of neural text degeneration}.
\newblock In \emph{8th International Conference on Learning Representations,
  {ICLR} 2020, Addis Ababa, Ethiopia, April 26-30, 2020}. OpenReview.net.

\bibitem[{Huang et~al.(2023)Huang, Yu, Ma, Zhong, Feng, Wang, Chen, Peng, Feng,
  Qin, and Liu}]{hall_survey_huang2023}
Lei Huang, Weijiang Yu, Weitao Ma, Weihong Zhong, Zhangyin Feng, Haotian Wang,
  Qianglong Chen, Weihua Peng, Xiaocheng Feng, Bing Qin, and Ting Liu. 2023.
\newblock \href {https://doi.org/10.48550/ARXIV.2311.05232} {A survey on
  hallucination in large language models: Principles, taxonomy, challenges, and
  open questions}.
\newblock \emph{CoRR}, abs/2311.05232.

\bibitem[{Jiang et~al.(2024)Jiang, Sablayrolles, Roux, Mensch, Savary, Bamford,
  Chaplot, Casas, Hanna, Bressand et~al.}]{jiang2024mixtral}
Albert~Q Jiang, Alexandre Sablayrolles, Antoine Roux, Arthur Mensch, Blanche
  Savary, Chris Bamford, Devendra~Singh Chaplot, Diego de~las Casas, Emma~Bou
  Hanna, Florian Bressand, et~al. 2024.
\newblock Mixtral of experts.
\newblock \emph{arXiv preprint arXiv:2401.04088}.

\bibitem[{Jin et~al.(2024)Jin, Zhang, Meng, Wang, and
  Tan}]{jin2024comprehensive}
Hanlei Jin, Yang Zhang, Dan Meng, Jun Wang, and Jinghua Tan. 2024.
\newblock A comprehensive survey on process-oriented automatic text
  summarization with exploration of llm-based methods.
\newblock \emph{arXiv preprint arXiv:2403.02901}.

\bibitem[{Kour et~al.(2022)Kour, Ackerman, Raz, Farchi, Carmeli, and
  Anaby{-}Tavor}]{measuring_the_measuring_tool_2022}
George Kour, Samuel Ackerman, Orna Raz, Eitan Farchi, Boaz Carmeli, and Ateret
  Anaby{-}Tavor. 2022.
\newblock \href {https://doi.org/10.48550/ARXIV.2211.16259} {Measuring the
  measuring tools: An automatic evaluation of semantic metrics for text
  corpora}.
\newblock \emph{CoRR}, abs/2211.16259.

\bibitem[{L{\'a}la et~al.(2023)L{\'a}la, O'Donoghue, Shtedritski, Cox,
  Rodriques, and White}]{lala2023paperqa}
Jakub L{\'a}la, Odhran O'Donoghue, Aleksandar Shtedritski, Sam Cox, Samuel~G
  Rodriques, and Andrew~D White. 2023.
\newblock Paperqa: Retrieval-augmented generative agent for scientific
  research.
\newblock \emph{arXiv preprint arXiv:2312.07559}.

\bibitem[{Lattimer et~al.(2023)Lattimer, Chen, Zhang, and
  Yang}]{ScreenEval_LattimerC0Y23}
Barrett~Martin Lattimer, Patrick Chen, Xinyuan Zhang, and Yi~Yang. 2023.
\newblock \href {https://doi.org/10.18653/V1/2023.EMNLP-MAIN.105} {Fast and
  accurate factual inconsistency detection over long documents}.
\newblock In \emph{Proceedings of the 2023 Conference on Empirical Methods in
  Natural Language Processing, {EMNLP} 2023, Singapore, December 6-10, 2023},
  pages 1691--1703. Association for Computational Linguistics.

\bibitem[{Li et~al.(2023)Li, Cheng, Zhao, Nie, and Wen}]{HaluEval_LiCZNW23}
Junyi Li, Xiaoxue Cheng, Xin Zhao, Jian{-}Yun Nie, and Ji{-}Rong Wen. 2023.
\newblock \href {https://aclanthology.org/2023.emnlp-main.397} {Halueval: {A}
  large-scale hallucination evaluation benchmark for large language models}.
\newblock In \emph{Proceedings of the 2023 Conference on Empirical Methods in
  Natural Language Processing, {EMNLP} 2023, Singapore, December 6-10, 2023},
  pages 6449--6464. Association for Computational Linguistics.

\bibitem[{Li et~al.(2018)Li, Kahou, Schulz, Michalski, Charlin, and
  Pal}]{redial}
Raymond Li, Samira~Ebrahimi Kahou, Hannes Schulz, Vincent Michalski, Laurent
  Charlin, and Chris Pal. 2018.
\newblock \href
  {https://proceedings.neurips.cc/paper/2018/hash/800de15c79c8d840f4e78d3af937d4d4-Abstract.html}
  {Towards deep conversational recommendations}.
\newblock In \emph{Advances in Neural Information Processing Systems 31: Annual
  Conference on Neural Information Processing Systems 2018, NeurIPS 2018,
  December 3-8, 2018, Montr{\'{e}}al, Canada}, pages 9748--9758.

\bibitem[{Liu et~al.(2024)Liu, Liu, Shi, Huang, Wang, Yang, and
  Zhang}]{liu2024exploring}
Fang Liu, Yang Liu, Lin Shi, Houkun Huang, Ruifeng Wang, Zhen Yang, and
  Li~Zhang. 2024.
\newblock Exploring and evaluating hallucinations in llm-powered code
  generation.
\newblock \emph{arXiv preprint arXiv:2404.00971}.

\bibitem[{Liu et~al.(2019)Liu, Ott, Goyal, Du, Joshi, Chen, Levy, Lewis,
  Zettlemoyer, and Stoyanov}]{roberta}
Yinhan Liu, Myle Ott, Naman Goyal, Jingfei Du, Mandar Joshi, Danqi Chen, Omer
  Levy, Mike Lewis, Luke Zettlemoyer, and Veselin Stoyanov. 2019.
\newblock \href {http://arxiv.org/abs/1907.11692} {Roberta: {A} robustly
  optimized {BERT} pretraining approach}.
\newblock \emph{CoRR}, abs/1907.11692.

\bibitem[{Moon et~al.(2019)Moon, Shah, Kumar, and Subba}]{opendialkg}
Seungwhan Moon, Pararth Shah, Anuj Kumar, and Rajen Subba. 2019.
\newblock \href {https://doi.org/10.18653/V1/P19-1081} {Opendialkg: Explainable
  conversational reasoning with attention-based walks over knowledge graphs}.
\newblock In \emph{Proceedings of the 57th Conference of the Association for
  Computational Linguistics, {ACL} 2019, Florence, Italy, July 28- August 2,
  2019, Volume 1: Long Papers}, pages 845--854. Association for Computational
  Linguistics.

\bibitem[{Muhlgay et~al.(2024)Muhlgay, Ram, Magar, Levine, Ratner, Belinkov,
  Abend, Leyton{-}Brown, Shashua, and Shoham}]{Factor_MuhlgayRMLRBALSS24}
Dor Muhlgay, Ori Ram, Inbal Magar, Yoav Levine, Nir Ratner, Yonatan Belinkov,
  Omri Abend, Kevin Leyton{-}Brown, Amnon Shashua, and Yoav Shoham. 2024.
\newblock \href {https://aclanthology.org/2024.eacl-long.4} {Generating
  benchmarks for factuality evaluation of language models}.
\newblock In \emph{Proceedings of the 18th Conference of the European Chapter
  of the Association for Computational Linguistics, {EACL} 2024 - Volume 1:
  Long Papers, St. Julian's, Malta, March 17-22, 2024}, pages 49--66.
  Association for Computational Linguistics.

\bibitem[{Ouyang et~al.(2022)Ouyang, Wu, Jiang, Almeida, Wainwright, Mishkin,
  Zhang, Agarwal, Slama, Ray, Schulman, Hilton, Kelton, Miller, Simens, Askell,
  Welinder, Christiano, Leike, and Lowe}]{instructGPT}
Long Ouyang, Jeffrey Wu, Xu~Jiang, Diogo Almeida, Carroll~L. Wainwright, Pamela
  Mishkin, Chong Zhang, Sandhini Agarwal, Katarina Slama, Alex Ray, John
  Schulman, Jacob Hilton, Fraser Kelton, Luke Miller, Maddie Simens, Amanda
  Askell, Peter Welinder, Paul~F. Christiano, Jan Leike, and Ryan Lowe. 2022.
\newblock \href
  {http://papers.nips.cc/paper\_files/paper/2022/hash/b1efde53be364a73914f58805a001731-Abstract-Conference.html}
  {Training language models to follow instructions with human feedback}.
\newblock In \emph{Advances in Neural Information Processing Systems 35: Annual
  Conference on Neural Information Processing Systems 2022, NeurIPS 2022, New
  Orleans, LA, USA, November 28 - December 9, 2022}.

\bibitem[{Pal et~al.(2023)Pal, Umapathi, and Sankarasubbu}]{MedHaLTPalUS23}
Ankit Pal, Logesh~Kumar Umapathi, and Malaikannan Sankarasubbu. 2023.
\newblock \href {https://aclanthology.org/2023.conll-1.21} {Med-halt: Medical
  domain hallucination test for large language models}.
\newblock In \emph{Proceedings of the 27th Conference on Computational Natural
  Language Learning, CoNLL 2023, Singapore, December 6-7, 2023}, pages
  314--334. Association for Computational Linguistics.

\bibitem[{Penedo et~al.(2023)Penedo, Malartic, Hesslow, Cojocaru, Alobeidli,
  Cappelli, Pannier, Almazrouei, and Launay}]{falcon}
Guilherme Penedo, Quentin Malartic, Daniel Hesslow, Ruxandra Cojocaru, Hamza
  Alobeidli, Alessandro Cappelli, Baptiste Pannier, Ebtesam Almazrouei, and
  Julien Launay. 2023.
\newblock \href
  {http://papers.nips.cc/paper\_files/paper/2023/hash/fa3ed726cc5073b9c31e3e49a807789c-Abstract-Datasets\_and\_Benchmarks.html}
  {The refinedweb dataset for falcon {LLM:} outperforming curated corpora with
  web data only}.
\newblock In \emph{Advances in Neural Information Processing Systems 36: Annual
  Conference on Neural Information Processing Systems 2023, NeurIPS 2023, New
  Orleans, LA, USA, December 10 - 16, 2023}.

\bibitem[{Peng et~al.(2023)Peng, Zhang, and
  Shang}]{LLM_generation_attribute_manipulation}
Letian Peng, Yuwei Zhang, and Jingbo Shang. 2023.
\newblock \href {https://doi.org/10.48550/ARXIV.2307.07099} {Generating
  efficient training data via llm-based attribute manipulation}.
\newblock \emph{CoRR}, abs/2307.07099.

\bibitem[{Singhal et~al.(2023)Singhal, Tu, Gottweis, Sayres, Wulczyn, Hou,
  Clark, Pfohl, Cole{-}Lewis, Neal, Schaekermann, Wang, Amin, Lachgar,
  Mansfield, Prakash, Green, Dominowska, y~Arcas, Tomasev, Liu, Wong, Semturs,
  Mahdavi, Barral, Webster, Corrado, Matias, Azizi, Karthikesalingam, and
  Natarajan}]{med-palm2}
Karan Singhal, Tao Tu, Juraj Gottweis, Rory Sayres, Ellery Wulczyn, Le~Hou,
  Kevin Clark, Stephen Pfohl, Heather Cole{-}Lewis, Darlene Neal, Mike
  Schaekermann, Amy Wang, Mohamed Amin, Sami Lachgar, Philip~Andrew Mansfield,
  Sushant Prakash, Bradley Green, Ewa Dominowska, Blaise~Ag{\"{u}}era y~Arcas,
  Nenad Tomasev, Yun Liu, Renee Wong, Christopher Semturs, S.~Sara Mahdavi,
  Joelle~K. Barral, Dale~R. Webster, Gregory~S. Corrado, Yossi Matias,
  Shekoofeh Azizi, Alan Karthikesalingam, and Vivek Natarajan. 2023.
\newblock \href {https://doi.org/10.48550/ARXIV.2305.09617} {Towards
  expert-level medical question answering with large language models}.
\newblock \emph{CoRR}, abs/2305.09617.

\bibitem[{Touvron et~al.(2023)Touvron, Martin, Stone, Albert, Almahairi,
  Babaei, Bashlykov, Batra, Bhargava, Bhosale, Bikel, Blecher, Canton{-}Ferrer,
  Chen, Cucurull, Esiobu, Fernandes, Fu, Fu, Fuller, Gao, Goswami, Goyal,
  Hartshorn, Hosseini, Hou, Inan, Kardas, Kerkez, Khabsa, Kloumann, Korenev,
  Koura, Lachaux, Lavril, Lee, Liskovich, Lu, Mao, Martinet, Mihaylov, Mishra,
  Molybog, Nie, Poulton, Reizenstein, Rungta, Saladi, Schelten, Silva, Smith,
  Subramanian, Tan, Tang, Taylor, Williams, Kuan, Xu, Yan, Zarov, Zhang, Fan,
  Kambadur, Narang, Rodriguez, Stojnic, Edunov, and Scialom}]{llama2}
Hugo Touvron, Louis Martin, Kevin Stone, Peter Albert, Amjad Almahairi, Yasmine
  Babaei, Nikolay Bashlykov, Soumya Batra, Prajjwal Bhargava, Shruti Bhosale,
  Dan Bikel, Lukas Blecher, Cristian Canton{-}Ferrer, Moya Chen, Guillem
  Cucurull, David Esiobu, Jude Fernandes, Jeremy Fu, Wenyin Fu, Brian Fuller,
  Cynthia Gao, Vedanuj Goswami, Naman Goyal, Anthony Hartshorn, Saghar
  Hosseini, Rui Hou, Hakan Inan, Marcin Kardas, Viktor Kerkez, Madian Khabsa,
  Isabel Kloumann, Artem Korenev, Punit~Singh Koura, Marie{-}Anne Lachaux,
  Thibaut Lavril, Jenya Lee, Diana Liskovich, Yinghai Lu, Yuning Mao, Xavier
  Martinet, Todor Mihaylov, Pushkar Mishra, Igor Molybog, Yixin Nie, Andrew
  Poulton, Jeremy Reizenstein, Rashi Rungta, Kalyan Saladi, Alan Schelten, Ruan
  Silva, Eric~Michael Smith, Ranjan Subramanian, Xiaoqing~Ellen Tan, Binh Tang,
  Ross Taylor, Adina Williams, Jian~Xiang Kuan, Puxin Xu, Zheng Yan, Iliyan
  Zarov, Yuchen Zhang, Angela Fan, Melanie Kambadur, Sharan Narang,
  Aur{\'{e}}lien Rodriguez, Robert Stojnic, Sergey Edunov, and Thomas Scialom.
  2023.
\newblock \href {https://doi.org/10.48550/ARXIV.2307.09288} {Llama 2: Open
  foundation and fine-tuned chat models}.
\newblock \emph{CoRR}, abs/2307.09288.

\bibitem[{Wang et~al.(2023)Wang, Li, Chen, Zhu, Lin, Cao, Liu, Liu, and
  Sui}]{llm_not_fair_evaluators}
Peiyi Wang, Lei Li, Liang Chen, Dawei Zhu, Binghuai Lin, Yunbo Cao, Qi~Liu,
  Tianyu Liu, and Zhifang Sui. 2023.
\newblock \href {https://doi.org/10.48550/ARXIV.2305.17926} {Large language
  models are not fair evaluators}.
\newblock \emph{CoRR}, abs/2305.17926.

\bibitem[{Wei et~al.(2022)Wei, Wang, Schuurmans, Bosma, Ichter, Xia, Chi, Le,
  and Zhou}]{cot_wei}
Jason Wei, Xuezhi Wang, Dale Schuurmans, Maarten Bosma, Brian Ichter, Fei Xia,
  Ed~H. Chi, Quoc~V. Le, and Denny Zhou. 2022.
\newblock \href
  {http://papers.nips.cc/paper\_files/paper/2022/hash/9d5609613524ecf4f15af0f7b31abca4-Abstract-Conference.html}
  {Chain-of-thought prompting elicits reasoning in large language models}.
\newblock In \emph{Advances in Neural Information Processing Systems 35: Annual
  Conference on Neural Information Processing Systems 2022, NeurIPS 2022, New
  Orleans, LA, USA, November 28 - December 9, 2022}.

\bibitem[{Xie et~al.(2024)Xie, Aggarwal, and Ahmad}]{continual_pretraining}
Yong Xie, Karan Aggarwal, and Aitzaz Ahmad. 2024.
\newblock \href {https://doi.org/10.18653/V1/2024.FINDINGS-ACL.606} {Efficient
  continual pre-training for building domain specific large language models}.
\newblock In \emph{Findings of the Association for Computational Linguistics,
  {ACL} 2024, Bangkok, Thailand and virtual meeting, August 11-16, 2024}, pages
  10184--10201. Association for Computational Linguistics.

\bibitem[{Yang et~al.(2023)Yang, Sun, and Wan}]{PHD_YangS023}
Shiping Yang, Renliang Sun, and Xiaojun Wan. 2023.
\newblock \href {https://doi.org/10.18653/V1/2023.FINDINGS-EMNLP.256} {A new
  benchmark and reverse validation method for passage-level hallucination
  detection}.
\newblock In \emph{Findings of the Association for Computational Linguistics:
  {EMNLP} 2023, Singapore, December 6-10, 2023}, pages 3898--3908. Association
  for Computational Linguistics.

\bibitem[{Ye et~al.(2022)Ye, Gao, Wu, Feng, Yu, and Kong}]{PROGEN_YeG0F0K22}
Jiacheng Ye, Jiahui Gao, Zhiyong Wu, Jiangtao Feng, Tao Yu, and Lingpeng Kong.
  2022.
\newblock \href {https://doi.org/10.18653/V1/2022.FINDINGS-EMNLP.269} {Progen:
  Progressive zero-shot dataset generation via in-context feedback}.
\newblock In \emph{Findings of the Association for Computational Linguistics:
  {EMNLP} 2022, Abu Dhabi, United Arab Emirates, December 7-11, 2022}, pages
  3671--3683. Association for Computational Linguistics.

\bibitem[{Yeti{\c{s}}tiren et~al.(2023)Yeti{\c{s}}tiren, {\"O}zsoy, Ayerdem,
  and T{\"u}z{\"u}n}]{yeticstiren2023evaluating}
Burak Yeti{\c{s}}tiren, I{\c{s}}{\i}k {\"O}zsoy, Miray Ayerdem, and Eray
  T{\"u}z{\"u}n. 2023.
\newblock Evaluating the code quality of ai-assisted code generation tools: An
  empirical study on github copilot, amazon codewhisperer, and chatgpt.
\newblock \emph{arXiv preprint arXiv:2304.10778}.

\bibitem[{Yu et~al.(2023)Yu, Zhuang, Zhang, Meng, Ratner, Krishna, Shen, and
  Zhang}]{YuZZMRKSZ23}
Yue Yu, Yuchen Zhuang, Jieyu Zhang, Yu~Meng, Alexander~J. Ratner, Ranjay
  Krishna, Jiaming Shen, and Chao Zhang. 2023.
\newblock \href
  {http://papers.nips.cc/paper\_files/paper/2023/hash/ae9500c4f5607caf2eff033c67daa9d7-Abstract-Datasets\_and\_Benchmarks.html}
  {Large language model as attributed training data generator: {A} tale of
  diversity and bias}.
\newblock In \emph{Advances in Neural Information Processing Systems 36: Annual
  Conference on Neural Information Processing Systems 2023, NeurIPS 2023, New
  Orleans, LA, USA, December 10 - 16, 2023}.

\bibitem[{Zhang et~al.(2023)Zhang, Li, Cui, Cai, Liu, Fu, Huang, Zhao, Zhang,
  Chen et~al.}]{zhang2023siren}
Yue Zhang, Yafu Li, Leyang Cui, Deng Cai, Lemao Liu, Tingchen Fu, Xinting
  Huang, Enbo Zhao, Yu~Zhang, Yulong Chen, et~al. 2023.
\newblock Siren's song in the ai ocean: a survey on hallucination in large
  language models.
\newblock \emph{arXiv preprint arXiv:2309.01219}.

\end{thebibliography}
